\documentclass[pdflatex,sn-mathphys-num,iicol]{sn-jnl}

\usepackage{graphicx}%
\usepackage{multirow}%
\usepackage{amsmath,amssymb,amsfonts}%
\usepackage{amsthm}%
\usepackage{mathrsfs}%
\usepackage[title]{appendix}%
\usepackage{xcolor}%
\usepackage{textcomp}%
\usepackage{manyfoot}%
\usepackage{booktabs}%
\usepackage{algorithm}%
\usepackage{algorithmicx}%
\usepackage{algpseudocode}%
\usepackage{listings}%

\algrenewcommand\algorithmicrequire{\textbf{Input:}}
\algrenewcommand\algorithmicensure{\textbf{Output:}}

\theoremstyle{thmstyleone}%
\theoremstyle{thmstyletwo}%

\theoremstyle{thmstylethree}%

\begin{document}
\title{From Vision To Language through Graph of Events in Space and Time: An Explainable Self-supervised Approach}


\author[1,2]{\fnm{Mihai} \sur{Masala}}\email{mihaimasala@gmail.com}

\author*[1,2]{\fnm{Marius} \sur{Leordeanu}}\email{leordeanu@gmail.com}

\affil*[1]{\orgname{Institute of Mathematics of the Romanian Academy}, \orgaddress{\street{21 Calea Grivitei Street}, \city{Bucharest}, \postcode{010702}, \country{Romania}}}

\affil[2]{\orgname{National University of Science and Technology POLITEHNICA Bucharest}, \orgaddress{\street{313 Splaiul Independentei}, \city{Bucharest}, \postcode{060042}, \country{Romania}}}


\abstract{The task of describing video content in natural language is commonly referred to as video captioning. Unlike conventional video captions, which are typically brief and widely available, long-form paragraph descriptions in natural language are scarce. This limitation of current datasets is due, on one hand, to the expensive human manual annotation required and, on the other, to the highly challenging task of explaining the language formation process from the perspective of the underlying story, as a complex system of interconnected events in space and time. Through a thorough analysis of recently published methods and available datasets, we identify a general lack of published resources dedicated to the problem of describing videos in complex language, beyond the level of descriptions in the form of enumerations of simple captions. Furthermore, while state-of-the-art methods produce impressive results on the task of generating shorter captions from videos by direct end-to-end learning between the videos and text, the problem of explaining the relationship between vision and language is still beyond our reach. In this work, we propose a shared representation between vision and language, based on graphs of events in space and time, which can be obtained in an explainable and analytical way, to integrate and connect multiple vision tasks to produce the final natural language description. Moreover, we also demonstrate how our automated and explainable video description generation process can function as a fully automatic teacher to effectively train direct, end-to-end neural student pathways, within a self-supervised neuro-analytical system. We validate that our explainable neuro-analytical approach generates coherent, rich and relevant textual descriptions on videos collected from multiple varied datasets, using both standard evaluation metrics (e.g. METEOR, ROUGE, BertScore), human annotations and consensus from ensembles of state-of-the-art large Vision Language Models. We also validate the effectiveness of our self-supervised teacher-student approach in boosting the performance of end-to-end neural vision-to-language models.}

\keywords{vision-language, graph of events in space and time, multi-task vision-to-language, video description in natural language, video captioning, self-supervised learning, neuro-analytic methods, trustworthy vision and language, visual story}



\maketitle

\section{Introduction}
\label{sec:intro}

The task of describing the visual content of videos in natural language, commonly referred to as video captioning~\cite{wang2018reconstruction,liu2018sibnet,aafaq2019spatio,li2021value,lin2022swinbert,chen2023valor}, represents a challenge for both the computer vision and natural language processing communities.

Although there is a plethora of powerful methods both from the field of video understanding (e.g. object detection and tracking~\cite{wang2023yolov7}, semantic segmentation~\cite{zou2023segment,cheng2021mask2former} and action recognition~\cite{tong2022videomae, wang2023videomae}), and from the area of
natural language processing (with LLMs such as GPT~\cite{brown2020language}), we are still far from understanding the bridge between vision to language. Consequently, we still do not have methods to describe in both rich and trustworthy natural language the content of videos. Although there are recent methods that learn from shared embeddings between vision and language~\cite{radford2021learning, li2022blip,zhai2023sigmoid}, they function as a black box and suffer from loss of information and lack of generalization. That is because images and videos are much richer than their short captions, so a lot of video content is lost. Moreover, the direct mapping between videos and their captions, without any thorough attempt to "understand" and "explain" the story behind the video, which unfolds in space and time, seems more like a look-up table, being strongly dependent on the particular training dataset, with its particular distribution and bias. 

Except for very recent commercial large Visual Language Models (VLMs), published deep learning methods trained for video description are only able to produce short and superficial textual video descriptions (video captions). They are rather close to video classification (where a video could belong to a finite number of classes) than to describing them in rich and free natural language, with its infinitely many forms for capturing a given video content. Moreover, such models strongly suffer from overfitting. Given a test video from an unseen data distribution, the quality and accuracy of the description drops, and our evaluations confirm that.

On the other hand, recent large VLMs demonstrate impressive results, with rich and fluent generated video descriptions.
However, we argue that current large VLMs are still heavily dependent on training data, generating interpolations in the high-level space of previously seen human-generated descriptions, rather than grounding their language output into a solid "explanation", that relates events and their relationships in the physical spatio-temporal world.
We also argue that such "spatio-temporal grounding" is needed for achieving generalization. Moreover, they only keep a few sampled frames for a video, thus missing from the start important information coming from frame-to-frame motion and key events. Consequently, their fluent and rich output often feels like an "hallucination" that cannot be trusted. On the website dedicated to this work\footnote{\url{https://sites.google.com/view/gest-project}}, we present such cases of hallucinations, produced by current state-of-the-art large VLMs.

We observed many such hallucinations, in all commercial state-of-the-art large VLM models we tested.  
However, they most often perform remarkably well, which could be due to their huge (and undisclosed) training set, besides their large and complex structure (also undisclosed).
It would not be surprising if these commercial models have been trained on the exact same videos that we test them on. Being commercial products (e.g. GPT, Claude or Gemini) and not fully open about their models and training data, we cannot consider them as scientific methods to fairly compare against. However, we could use them as automatic annotators or evaluators, given their high-quality output. Moreover, when used as an ensemble, their combined output suffers less from lack of generalization and is in better agreement with human annotations, as clearly shown in our experiments (Section~\ref{sec:human_vs_metrics}).

In conclusion, we are still lacking a published formal and explicit way to relate vision and language, to extract and explain the story that unfolds in space and time, at the level of meaning and how events relate, behind the front-end forms of video and language. 

In this work, we address this challenge and propose a general approach that is grounded in the visual understanding of actors and objects, their actions and interactions, their scene context, 3D proprieties and movements, to form a graph of events in space and time (GEST) - which we propose as a shared explicit representation of stories, between vision and language (Section~\ref{sec:gest}). From GEST we then cross into the realm of language, by procedurally transforming GEST into text, to produce a description in proto-language, which is fed into a LLM module, to obtain the final refined fluent language form (Section~\ref{sec:method}). Our model also contains skip neural connections, acting as student paths, which could learn in a self-supervised way, from the explainable multi-task pathway, acting as the unsupervised teacher.

Vision-language is a complex domain, which is why our approach is fundamentally multi-task, such that each task is a step, a module that helps in establishing the bridge from vision to language. Our methodology establishes relationships between multiple tasks across both space and time, ensuring that the narrative structure of the description aligns with the sequence of events as they unfold. The outcome of this approach is a Graph of Events in Space and Time - GEST, published in its initial version at ICCV 2023~\cite{masala2023explaining}. GEST effectively captures and represents the dynamic progression and interactions of these events. GEST represents stories, with actors performing actions and creating events that relate to each other, at different levels of abstraction, producing a hierarchy such that every event could be broken up into a lower-level GEST and every GEST could become an event at a higher-level. Such a representation can naturally express both vision and language and we provide extensive experiments to prove our point. Our experiments (Sections~\ref{sec:experimental_chapter} and~\ref{sec:eval_chapter}), with extensive evaluations over several datasets from both humans and state-of-the-art large VLMs, also reveal several limitations of current datasets. They are rather simple and mostly suited for short caption descriptions, with classic text similarity metrics - which compare texts mostly at the level of form, not meaning.
 
\begin{figure*}
\includegraphics[width=1.0\linewidth]{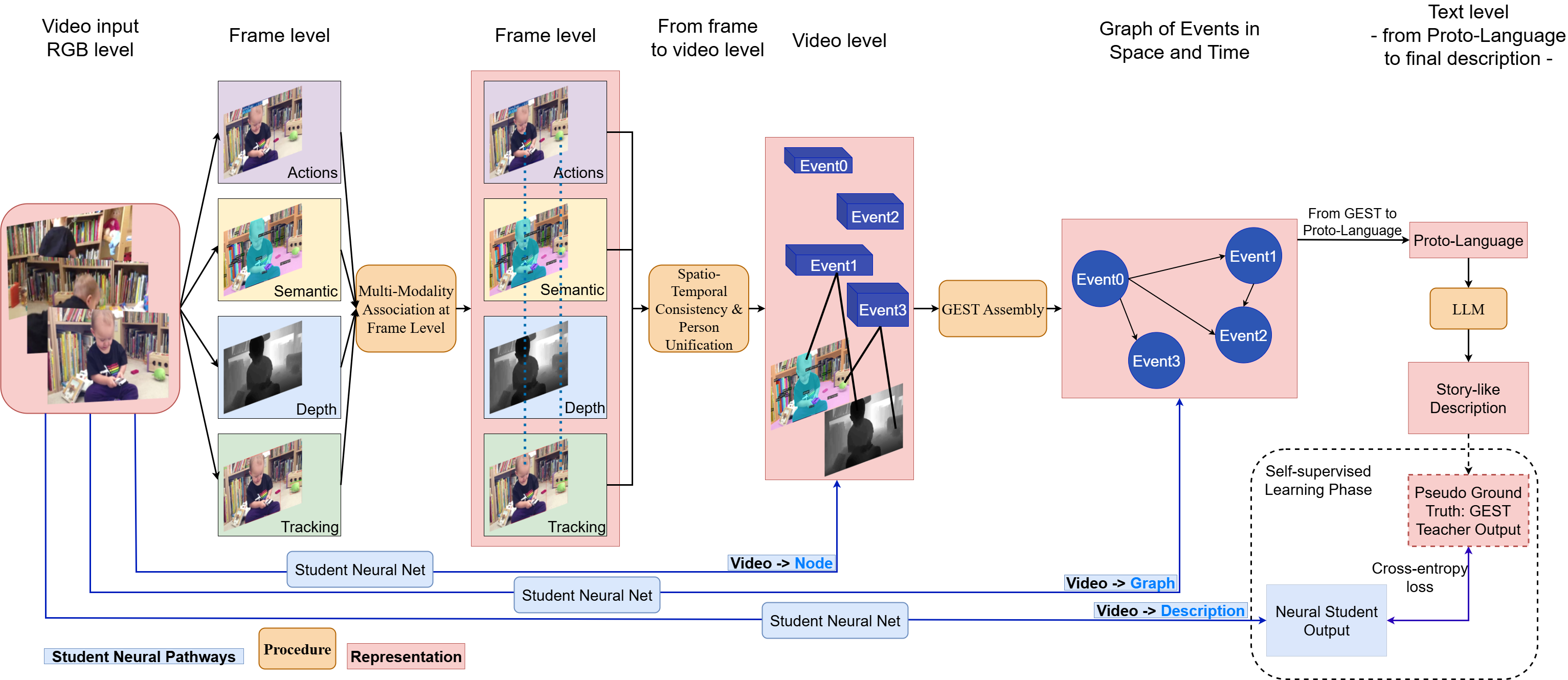}
\caption{A modular, algorithmic step-by-step outline of our proposed approach. Starting from the video we harness multiple tasks (i.e., action detection, semantic segmentation, depth estimation and tracking), followed by a second step of multi-modality association. While we still have frame-level information, at this point information is aggregated and correlated across modalities. A third step follows, in which we ensure space and time consistency and person unification,  to generate a list of events containing the actions, unique person ids, location, objects and start and end frame, transitioning to video level representation. At the fourth step, this list is then directly converted into GEST nodes and GEST edges are created to yield the final GEST graph. In the self-supervised learning module, we demonstrate how neural student networks (skip-like connections shown in blue, below) can learn from their corresponding GEST teacher path (or sub-path, above). While all such student pathways can learn from their respective GEST teacher paths, in our experiments we focused on the longest connection from original video input to final text output, as illustrated in the self-supervised learning box in the figure.}
\label{fig:architecture}
\end{figure*}

An overview of our proposed approach is presented in Figure~\ref{fig:architecture}. Our contribution can be broken up into the following main parts:
\begin{itemize}
    \item By linking multiple tasks across space and time, we automatically generate Graphs of Events in Space and Time $-$ an extension of traditional scene graphs~\cite{johnson2015image} into the spatiotemporal domain. This approach formally represents the concept of a story in which events, driven by actors, interact.  
    \item We devise a method to generate an intermediate language description (i.e., proto-language) for any such GEST. This proto-language is subsequently refined using text-only large language models (LLMs) to produce accurate and semantically rich video descriptions.
    \item Our experimental results show that descriptions generated by our approach are preferred both by human evaluators and automated methods, including text similarity metrics and state-of-the-art large VLMs. 
    \item We show that even datasets traditionally regarded as extensive and complex are better suited for the task of short-form video captioning rather than generating rich, story-style video descriptions. There is a clear need for datasets appropriate for such rich narrative descriptions, to move away from the rather simplistic video captioning and develop next-generation video to language models.
    \item Furthermore, we show that standard text similarity metrics exhibit lower alignment with human preferences and fail to capture subtle nuances effectively. We show that a jury of large VLMs is a more reliable proxy for human preferences.
    \item We prove the quality and effectiveness of the generated data, in a self-supervised framework for pretraining and increasing the quality and performance of end-to-end vision-to-language models.
    \item Finally, we will make all resources, from code to evaluation results, publicly available\footnote{\url{https://sites.google.com/view/gest-project}} to support further study and research advancements. 
\end{itemize}

\section{Related Work}
Up until recently, most video-to-text models were based on the encoder decoder architecture, using mostly CNNs for encoding the video frames and LSTMs to generate the textual description~\cite{wang2018reconstruction,liu2018sibnet}. Research~\cite{li2021value} has been focused on probing different video representations such as ResNet~\cite{he2016deep}, C3D~\cite{hara2018can} and S3D~\cite{miech2020end} or CLIP-ViT~\cite{sun2023eva}, for improving video captioning quality. 

Dosovitskiy~\cite{dosovitskiy2020image} showed that the Transformer architecture, which has been initially developed for machine translation, can also be applied in computer vision tasks, outperforming CNNs in image classification tasks. From then on, Transformers have been successfully applied in a broad range of Computer Vision tasks including tasks performed on videos: action recognition~\cite{liu2022video}, video captioning~\cite{lin2022swinbert} or even multi-modal (vision and language) learning~\cite{fu2023empirical,chen2023valor,chen2023cosa}. VALOR~\cite{chen2023valor} uses three separate encoders for video, audio and text modalities and a single decoder for multi-modal conditional text generation. This architecture is pretrained on 1M audible videos with human annotated audiovisual captions, using multi-modal alignment and multi-modal captioning tasks. PDVC~\cite{wang2021end} frame the dense caption generation as a set prediction tasks with competitive results, compared to previous approaches based on the two-stage “localize-then-describe” framework.

Unified vision and language foundational models are either trained using both images and videos simultaneously~\cite{alayrac2022flamingo} or use a two-stage approach~\cite{wang2021simvlm, yu2022coca} in which the first stage contains image-text pairs, followed by a second stage in which video-text pairs are added. This two-stage approach has the advantage of faster training, models can be scaled up easier, and data is more freely available. VAST~\cite{chen2023vast} is a unified foundational model across three modalities: video, audio and text. To alleviate the limited scale and quality of video-text training data, COSA~\cite{chen2023cosa} converts existing image-text data into long-form video data. Then an architecture based on ViT~\cite{dosovitskiy2020image} and BERT~\cite{devlin2019bertpretrainingdeepbidirectional} is trained on this new long-form data. GIT~\cite{wang2022git} is a unified vision-language model with a very simple architecture consisting of a single image encoder and a text decoder, trained with the standard language modeling task. mPLUG-2~\cite{Xu2023mPLUG2AM} builds multi-modal foundational models using separate modules including video encoder, text encoder, image encoder followed by universal layers, a multi-modal fusion module and finally a decoder module. It is important to note that the methods mentioned above are rooted in the classical and limited image-only domain, in which time is not given the deserved importance. By sampling frames, fundamentally ignoring the idea of movement, we could very well miss important events. Furthermore, training large models as black boxes misses the problem they tackle, which is that of understanding how things related and develop from the physical world, vision and senses, to language.

VidIL~\cite{wang2022language} is one of the first methods that decouples the vision and language parts from the video captioning task. It uses CLIP~\cite{radford2021learning} and BLIP~\cite{li2022blip} to extract image descriptors (i.e. image captions, events, attributes and objects) from uniformly sampled frames, descriptors that are then fed to a text-only LLM. Our proposed approach is similar in principle, but we bring a much stronger, grounded and more complex methodology to extract video descriptors from multiple modalities. Besides processing every individual frame from the video stream (as opposed to sampling four frames in case of VidIL), we aggregate multi-task information across space and time, leading to complex spatial and temporal relations. The temporal information in VidIL is obtained by simply concatenating the resulting descriptions and incorporating temporal "tokens" (i.e., "first", "then", "finally"), modeling only the "next" temporal relation.

While scene graphs~\cite{johnson2015image} were originally proposed and used for image retrieval, they have become more and more popular, currently being used in a variety of tasks including image captioning~\cite{wang2020storytelling,chen2020say,xu2019scene}, image generation~\cite{johnson2018image,li2019pastegan,dhamo2020semantic} or visual question answering~\cite{ghosh2019generating,zhang2019empirical}. Video scene graphs~\cite{shang2017video} are a natural extension of scene graphs in the time domain. Generating such video scene graphs from a video include methods based on image scene graph generation~\cite{ji2020action} or methods based on video encoder followed by relationship extraction~\cite{zheng2022vrdformer}. Specialized versions such as panoptic~\cite{yang2023panoptic} and egocentric~\cite{rodin2024action} scene graph generation have been developed. As scene graphs model semantic relationships between objects in images or videos, with Graph of Events in Space and Time we are more interested in events, interactions between actors and between actors and objects. GEST framework is not limited to $(subject,predicate,object)$ triplets (e.g. GEST can capture complex, multi-person, multi-object interactions) and explicitly models events through space and time.

Long, complex textual descriptions of videos also appear in the context of multi-sentence description~\cite{shin2016beyond,park2019adversarial,park2020identity}.~\cite{ziaeetabar2024multi} proposes a multi-sentence description method for complex manipulation action videos that leverages YOLO v3~\cite{redmon2018yolov3}, OpenPose~\cite{cao2017realtime} and depth estimation. For this specialized task, a set of predefined spatial relations (e.g., around, inside) and dynamic spatial relations (e.g., moving apart, halting together) are used. Video ReCap~\cite{islam2024video} is a hierarchical architecture of three levels that is capable of describing hour-long videos. Starting from standard video captioning on videos of a few seconds, is followed by segment descriptions that are generated from sparsely sampled video features and the previously generated clip captions of a particular segment. Similarly, the final stage uses video feature and segment descriptions sampled from the entire video.

\citet{nadeem2024narrativebridge} make use of LLMs to augment existing captions with explicit cause-effect temporal relationships in video descriptions to build a novel Causal-Temporal Narrative (CTN) captions benchmark dataset. A Cause-Effect Network, a two-stage architecture based on separate branches for cause and effect is devised to generate captions with causal-temporal narrative. Similarly, \citet{yu2018fine} augment 2000 videos from NBA games with fine-grained captions or Sports Narratives. A three branch architecture, consisting of a spatio-temporal entity localization and role discovering for team partition and player localization, an action modeling sub-network that used a human skeleton motion description module and a sub-network for modeling interactions among players is built for this task. Features from the three branches are then fused and fed to a hierarchically recurrent decoder that generated the rich, story-like description.

\section{Graph of Events in Space and Time - GEST}
\label{sec:gest}

In this section we present the more theoretical aspects of GEST, which in initial form was introduced by Masala et al.~\cite{masala2023explaining}.

At its core, a Graph of Events in Space and Time (GEST) serves as a framework for representing stories. Interactions between events across space and time alter state of the world, can trigger or cause other events and in turn cause further changes. As a result, these events and their interactions serve as the building blocks of GEST.

In GEST nodes represent events that could go from simple actions (e.g. opening a door) to complex, high-level events (e.g. a political revolution), in terms of spatio-temporal extent, scale and semantics. They are usually confined to a specific time period (e.g. a precise millisecond or whole year) and space region (e.g. a certain room or entire country). Events could exist at different levels of semantics, ranging from simple physical contact (e.g. ``I touch the door handle'') to profoundly semantic ones (e.g. ``the government has fallen'' or ``John fell in love with physics''). Even physical objects are also events (e.g. John's car is represented by the event ``John's car exists''). Generally, any space-time entity could be a GEST event.

Edges in GEST relate two events and can define any kind of interaction between them, from simple temporal ordering (e.g. ``the door opened'' after ``I touched the door handle'') to highly semantic (e.g. ``the revolution'' caused ``the fall of the government'', or ``Einstein's discovery'' inspired ``John to fall in love with physics''). Generally, any verb that relates two events or entities could be a GEST edge.

A Graph of Events in Space and Time essentially represents a story in space and time, which could be arbitrarily complex. Even simple events can be explained by a GEST, since all events can be broken down, at a sufficient level of detail, into simpler ones and their interactions (e.g. ``I open the door'' becomes a complex GEST if we describe in detail the movements of the hand and the mechanical components involved). At the same time, any GEST could be seen as a single event at a higher semantic and spatio-temporal scale (e.g. ``a political revolution'' could be both a GEST and a single event). Collapsing graphs into nodes ($Event \Leftarrow GEST$) or expanding nodes into graphs ($GEST \Leftarrow Event$)
, gives GEST the ability to have many levels of depth, as needed for complex visual and linguistic stories.

\begin{figure}
    \centering
    \includegraphics[width=\columnwidth]{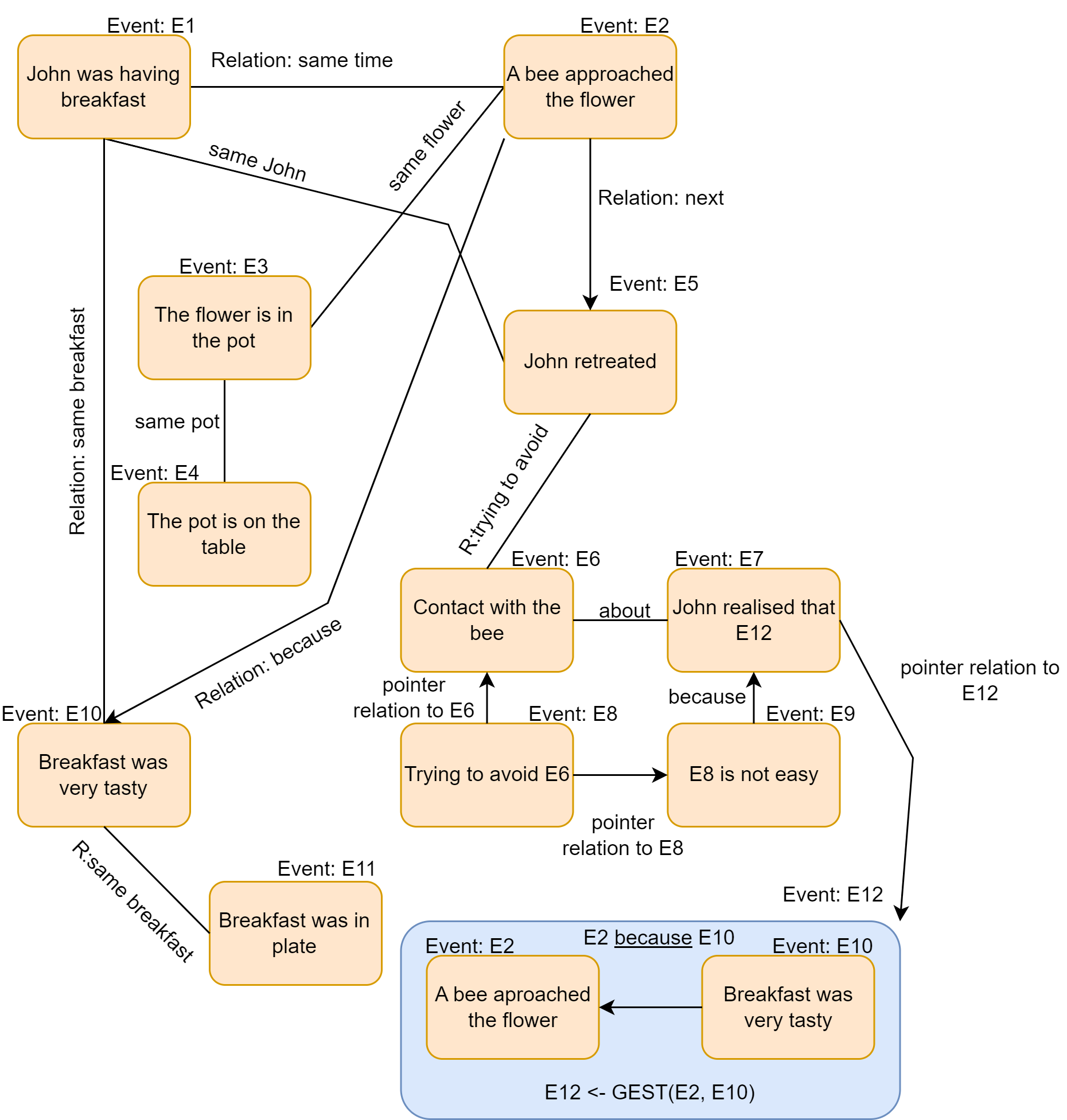}
    \caption{GEST graph explaining the following text: \textit{``John was having breakfast when a bee approached the flower in the pot on the table. Then he pulled back trying to avoid contact with the bee but he realized that it was not an easy attempt because she actually came because of the tasty food on his plate''.}}
    \label{fig:gest_example_complex}
\end{figure}

Going from a $GEST$ at a lower level to an event $E$ at a higher level ($E \Leftarrow GEST$) is reminiscent of how the attention mechanism is applied in Graph Neural Networks and Transformers \cite{vaswani2017attention}: the GEST graph acts as a function that aggregates information from nodes (events) $E_i$'s at level $k$ and builds a higher level GEST representation, which further becomes an event at the next level $k+1$:

\[ E_i^{(k+1)} \Leftarrow GEST(E_1^k, E_2^k, ..., E_n^k)\]

In Fig. \ref{fig:gest_example_complex} we present the proposed GEST representation applied to a specific text. In each event node, $E_i$, we encode an $action$, a list of $entities$ that are involved in the action, its $location$ and $timeframe$ and any additional $properties$. Note that an event can contain references (pointers) to other events, which define relations of type ``same X'' (e.g. ``same breakfast''). We also exemplify how the GEST of two connected events can collapse into a single event.

\section{Method}
\label{sec:method}
To properly describe all kinds of videos ranging from simple to more complex (e.g., longer, with more actions and actors) we first have to analyze and understand what happens in a video. Furthermore, to ensure this process is explainable we decide to stray away from the current paradigm of sampling frames, processing, and feeding them into a model that builds an inner obfuscated numerical representation. Instead, we aim to harness the power and expressivity of Graphs of Events in Space and Time (GESTs). Therefore, our first goal is to understand the video, to build a pipeline that given an input video it automatically builds an associated GEST. Then, by reasoning over GEST we build an intermediate textual description in the form of a proto-language that is then converted to natural language description.

\begin{figure*}
\includegraphics[width=1.0\linewidth]{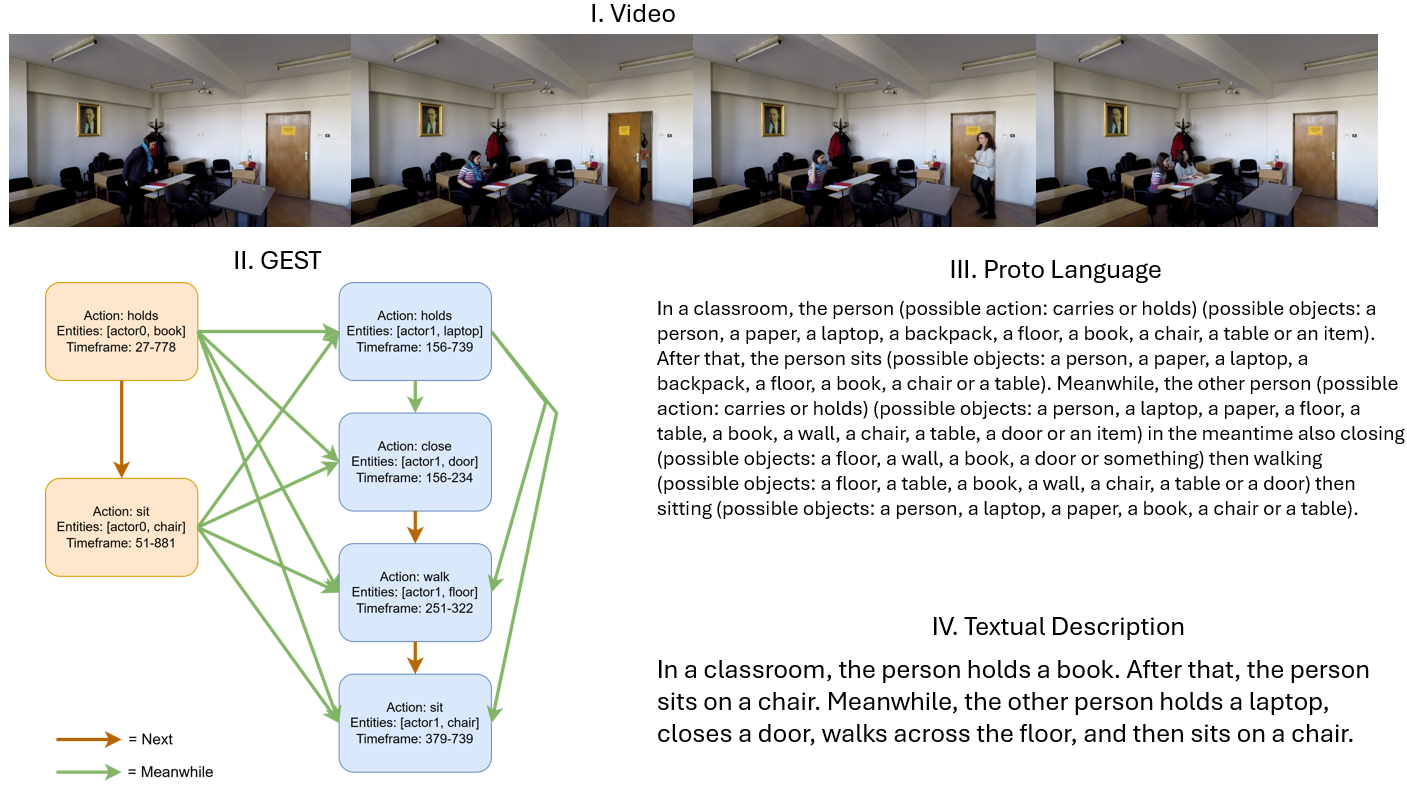}
\caption{A complete example of our proposed pipeline. Starting from the video (I), we we automatically build the associated GEST (II), capturing events and relations between events. From this graph, we build the proto-language (III) that is then fed to an LLM that generates the final textual description (IV).}
\label{fig:complete_example}
\end{figure*}

In summary, our framework consists of two main steps: I. building the Graph of Events in Space and Time by processing and understanding frame level information, followed by reasoning to get an integrated, global view and II. translating this understanding in a rich natural language description by reasoning over GEST via a two-step process. A detailed step-by-step outline is presented in Figure~\ref{fig:architecture}, while a complete example, starting from a video all the way to the final description is presented in  Figure~\ref{fig:complete_example}.

\subsection{Understanding the video - Building the GEST}

To construct an explicit representation of a video, we incorporate multiple tasks — primarily from the field of computer vision. In principle, for describing events in natural language, these tasks could also involve other sensory modalities, such as hearing, touch, and temperature. Returning to the problem at hand, we exploit existing sources of high-level image and video information, related to different tasks: action detection, object detection and tracking, semantic segmentation and depth information. For each frame in a given video, we first extract this information followed by a matching and aggregation step. The output of the action detector includes, for every action a bounding box of the person performing the action together with the name of the action and a confidence score. Starting from this bounding box, we aim to gather all the objects in the vicinity of the person, objects with which the actor could interact. First, the original bounding box is slightly enlarged to better capture the surroundings of the person, followed by finding all the objects that touch or intersect the new bounding box, based on information from the object detector and semantic segmentation. The list of objects is further filtered based on the intersection over union of the object and person bounding box with a fixed threshold, followed by depth-based filtering: we compute an average pixel-level depth for the person and the object, and if the depth difference between the person and the objects is between a set threshold, we consider the object close enough (both in "2D" based on intersection of bounding boxes and in "3D" based on depth) and we keep it in the list. All the objects that are not in the proximity of the person are discarded. Using this process, at this step we save for each action at each frame, information that includes the frame number, the person id (given at this point by the tracking model), the action name and confidence score as given the action detector, possibly involved objects and the bounding box of the person.

The next step is to aggregate and process frame-level information into global video-level data. The first thing we noticed was that the model used for tracking had slight inconsistencies (e.g., changing the assigned id for a person from one frame to another even though the person in question did not move) or certain blind spots (e.g., losing sight of a person for a couple of frames). Upon detecting the person again it assigns a new id, as if it was a different person. We solve these short-term inconsistencies by unifying two person ids if they appear close in time (less than 10 frames) and they overlap enough (higher than 0.6 intersection over union). Note that these thresholds were set empirically by manually verifying around 20-25 examples. 

\begin{figure}
\includegraphics[width=1.0\linewidth]{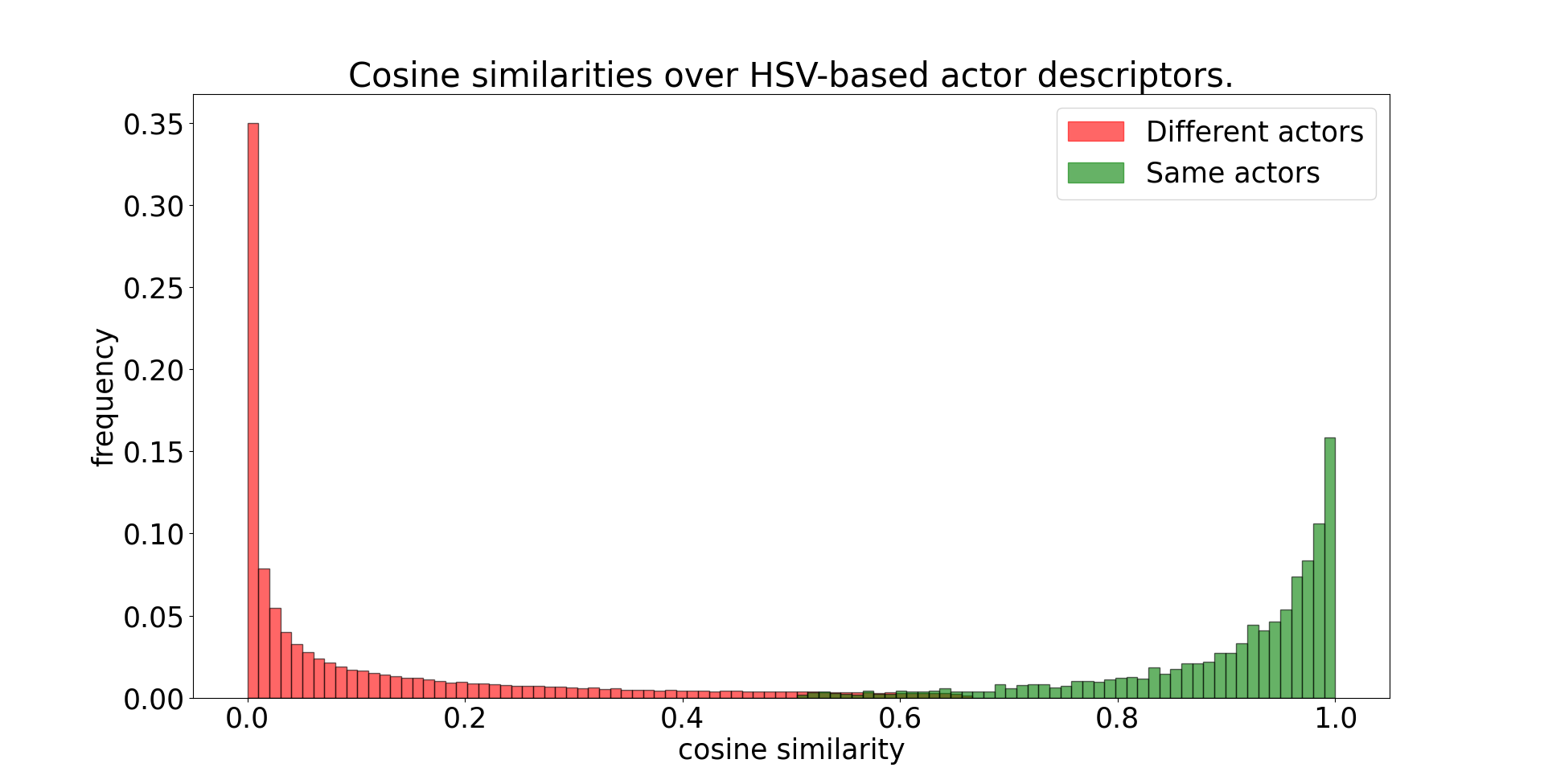}
\caption{Histogram of cosine similarities over HSV-based person descriptors on a small, curated set of videos. Based on these results, we consider cosine similarities of over 0.3 to be indicative of the same person.}
\label{fig:hist_poz_neg}
\end{figure}

The lack of consistency of the tracker manifests itself both in short-term and long-term inconsistencies. An example of long-form inconsistency is when a person exists the frame, either due to camera movement or the person moving, and then re-enters the frame at a later time and in a different position. The previous solution cannot work for this long-term inconsistency. Instead we are looking for a semantic-based solution that is powerful enough for person re-identification while being very fast. For each person detected, using representative frames (e.g., frames in which the person's masks has a high number of pixels), we efficiently compute an appearance feature vector based on the HSV histogram. For each pixel in the segmentation/mask of a person we bin the hue, saturation and value and linearize the resulting 3 dimensional space into a vector (see Algorithm~\ref{alg:person_descriptor}). Finally, we use cosine similarity to compare the resulting descriptors. If the highest similarity exceeds a set threshold, we unify them into a single entity. For finding a fair threshold, we generate positive (same actors) and negative (different actors) pairs of actors for 40 videos. Both positive and negative pairs are based on YOLO tracking ids that we further clean and curate to ensure their correctness, in the end being left with around 2500 positive pairs and 143k negative pairs. A histogram of cosine similarities is presented in Figure~\ref{fig:hist_poz_neg}. We empirically set the similarity threshold to a value of 0.3. Note that, while the linearized HSV histogram is simple and fast to compute, it reliably captures the appearance of a person within a video clip, as the distribution of colors for a given person is relatively stable, wearing the same clothing during the relatively short clip. While other features could have been extracted in order to establish the identity of a person from their segmentation mask, we found the HSV histogram to be simple, fast and reliable.

\begin{algorithm}
\caption{compute\_person\_descriptor}\label{alg:person_descriptor}
\begin{algorithmic}[1]
\Require $person\_data (pdata)$
\Ensure $descriptor (d)$
\State $masks = []$
\State $pixels = []$
\For{frame in $pdata.start\_frame$:$pdata.end\_frame$}
    \State \texttt{$masks.add(pdata.frame.mask)$}
    \State \texttt{$pixels.add(count(pdata.frame.mask))$}
\EndFor
\State $avg\_pxs = mean(pixels)$
\State $std\_pxs = std(pixels)$

\State $repr\_masks = []$
\For{$mask, pixels$ in $masks, pixels$}
    \If{$pixels \geq avg\_pxs+1.5*std\_pxs$}
        \State $repr\_masks.add(mask)$
    \EndIf
\EndFor

\State $d = zeros(20*10*5)$
\For{$repr\_mask$ in $repr\_masks$}
    \State $mask\_hsv = convert\_to\_hsv(repr\_mask)$
    \For{$pixel$ in $mask\_hsv$}
        \State $binH = bin(pixel.H, num\_bins=20)$
        \State $binS = bin(pixel.S, num\_bins=10)$
        \State $binV = bin(pixel.V, num\_bins=5)$
        \State $d[binH*50+binS*5+binV] += 1$
    \EndFor
\EndFor

\State \Return $d$
\end{algorithmic}
\end{algorithm}

The next step after person unification, is frame-based action filtering: based on empirically set thresholds, we filter our actions with confidence lower than 0.75 and for each frame keep only the two most confident actions. Then, to ensure a certain robustness, we implement a voting mechanism as follows: for each action in a frame we consider the previous five and the next five frames and if an action appears less than five times in this window of 11 frames, we discard it. This voting mechanism alleviates some of the inconsistencies of the action detector and person tracker and ensures a smoother action space. 

Armed with this rich frame-level information we proceed to build the video-level representation. The first step is aggregating actions that appear in consecutive frames in events by saving the start and end frame ids, possible objects involved (union over objects at each frame, keeping objects that appear at least in 10\% of frames between start and end frame) and bounding boxes. Finally, we perform an additional unification step in which we aim to detect cases in which we find events with the same actors and the same action that are close in time (e.g. one starts at frame 10 and ends at frame 120 while the second starts at frame 130 and ends at frame 250) but are considered two different events. As such, we unify such events, again to make the final event-space less fragmented and more coherent. 

At this moment we have a list of events and for each event have actors, objects, timeframe (start and end frame ids) and location (bounding boxes). The last step in this entire pipeline in building spatio-temporal relationships between events. As both temporal and spatial information is readily available for each event, this is a rather straightforward process: we build pairs of events and if they meet certain criteria we link them in space or time. For spatial relations between two events that have an overlap in time, for each such frame, we are interested in the two actions being close in space. Therefore we compute the ratio between the Euclidean distance of the centroids and the sum of the diagonals and if this ratio is lower than a certain threshold we consider that the two actions are related (i.e., close) in space. If this happens for more than 75\% of the overlapping frames we consider the events to be close in space and mark them accordingly (i.e. build an edge, a spatial relation in the graph between the two events). For temporal relations we follow a similar approach, we are checking pair of events, characterize three types of temporal relations: next, same time and meanwhile. 

This leaves us with an over-complete graph, as it contains an over-complete set of possible objects for each event. Better grounding and obtaining a concrete GEST can be obtained in a variety of ways including picking objects based on proximity to the person or by the "temporal" size (number of frames in which is close to the person). We solve this at a later stage in the pipeline, by allowing an LLM to pick the most probable object. For more details see the following section.

For action detection, we use VideoMAE~\cite{tong2022videomae} finetuned on AVA-Kinetics~\cite{li2020ava}. Object detection and tracking are performed using the YOLO pipeline~\cite{yolov8_ultralytics}, while semantic segmentation is performed using Mask2Former~\cite{cheng2021mask2former}. Finally, Marigold~\cite{ke2023repurposing} is used to compute depth estimation. Hyperparameters used for building the Graph of Events in Space and Time are presented in Table~\ref{tab:hyperparms}.

\begin{table}
  \centering
  \begin{tabular}{lcc}
    \toprule
    Description & Value\\
    \midrule
    Actions to keep at each frame & Top 2\\
    Action confidence threshold & $\geq 75\%$\\
    Distance between consecutive events to unify & $\leq 10\%$\\
    Frames left-right for window action filtering & $5$\\
    Voting threshold for window action filtering & $\geq 45\%$\\
    Actor-object interaction IOU threshold & $\geq 0.1$\\
    Depth threshold & $\le 25\%$\\
    Person re-id threshold & $\geq 0.3$\\
    \bottomrule
  \end{tabular}
  \caption{Hyperparameters used for generating GESTs.}
  \label{tab:hyperparms}
\end{table}

\subsection{Generating a natural language description}

Translating a GEST into a coherent, rich and natural language description is not a straight-forward task with multiple possibilities. In this work we adopt a two-stage approach that harnesses the power of existing text-based LLMs to build natural descriptions. The goal of the first step in our approach in to convert the graph into a sound (but maybe a bit rough around the edges) textual form, an initial description that we call "proto-Language". While this representation is valid and accurately depicts the information encoded in the graph, it lacks a certain naturalness, as it may sound too robotic/programmatic, lacking a more nuanced touch. Therefore, to obtain a more human-like description we use existing LLMs by feeding them with this proto-Language and prompting with the goal of rewriting the text to make it sound more natural.

\begin{algorithm}
\caption{build\_proto\_language}\label{alg:build_proto_language}
\begin{algorithmic}[1]
\Require $g = GEST$
\Ensure $descriptor (d)$
\State $done\_events = []$
\State $parent\_events = []$
\State $groups = []$
\State $events = sort(g.events, key=start\_frame)$
\For{event in $events$}
    \State $crt\_event = event$
    \State $crt\_group = [crt\_event]$
    \If{$all(event.links.type$=="Next") and $not(event.done)$}
        \State $parent\_events.add(crt\_event)$
        \State $groups.add(crt\_group)$
        \For{$ev$ in $crt\_group$}
            \State $en.done = True$
        \EndFor
        \State continue
    \EndIf

    \State $added\_new\_event=False$
    \State $linked\_events$ = $crt\_event.links$ if $same\_actor$
    
    \For{$linked\_event$ in $linked\_events$}
        \State $crt\_group.add(linked\_event)$
        \If{not $linked\_event.done$}
            \State $added\_new\_event=True$
        \EndIf
    \EndFor

    \If{not $added\_new\_event$}
        \If{$crt\_event.done$}
            \State continue
        \Else
            \State $filter(not(done), crt\_group)$
        \EndIf
    \EndIf
    \State $parent\_events.add(crt\_event)$
    \State $groups.add(crt\_group)$
    \For{$ev$ in $crt\_group$}
        \State $ev.done=True$
    \EndFor   
    \EndFor
    \State $descs=[]$ 
    \For{$group$ in $groups$}
        \If{$group$ is $first$}
            \State $conn = None$
        \Else
            \State $conn$ = $spatio\_temporal(group, prev\_group)$
        \EndIf
        \State $desc=describe\_group(group)$
        \State $descs.add(conn+desc)$
    \EndFor
\State \Return $descs$
\end{algorithmic}
\end{algorithm}

\begin{figure*}
\includegraphics[width=1.0\linewidth]{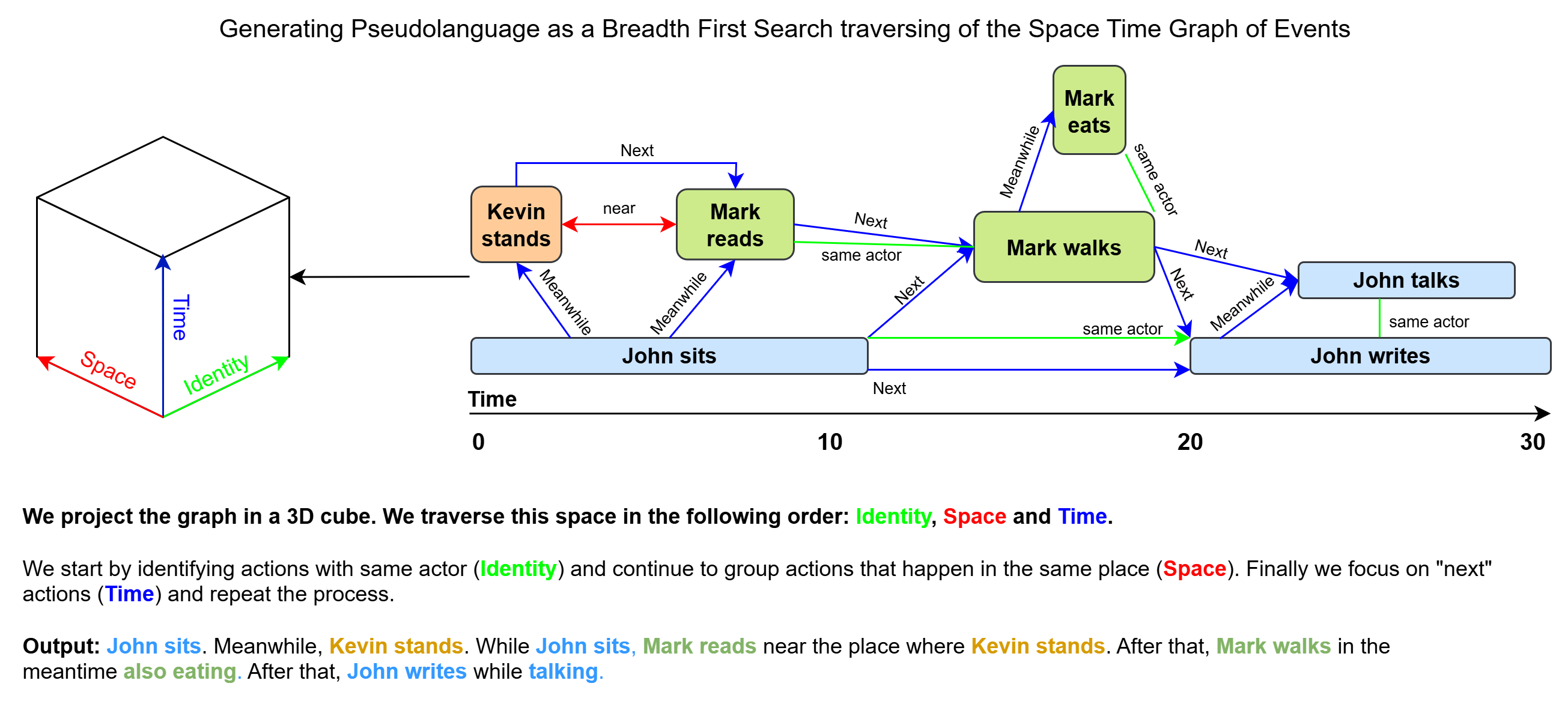}
\caption{Generating the proto-language via mapping the Graph of Events in Space and Time into a 3D volume and traversing this volume in Identity, Space and Time order.}
\label{fig:order}
\end{figure*}

The visual information is already converted and integrated into the GEST, but the question of how this graph can be effectively converted to an input to be consumed by an LLM still remains. The first step in this process involves a temporal sorting of the graph (by the start frame of each event; akin to a topological sort). If at each moment in time a single actor performs a single action, this is a rather straightforward process, with the results being a tree in space and time. With multiple actors and/or actions, this becomes more complex, with more than one possible representation. Our approach maps the graph of events in space and time into a 3D volume and we build the proto-language by breadth-first-search over this space. We aggregate chronologically sorted actions into higher-level groups of actions by actors. Each such group is then described in text, by describing each event using a simple grammar and taking into account the intra-group and inter-group spatial and temporal relations. A high-level overview of this algorithm is presented in Figure~\ref{fig:order} with a pseudo-code implementation presented in Algorithm~\ref{alg:build_proto_language}. Describing a single event involves describing the actor or actors (including objects) involved, the action performed, and spatial and temporal information if available. 

Crucially, we decide to not make a hard decision when selecting the possible objects involved in an event and to double down on the power of LLMs, feeding them with special instructions for selecting the most probable object in the given context. Therefore, when describing an event, we list all possible objects (as computed earlier)  and let the LLM pick the objects that are most probable to appear in the given context, with the power to pick a new object that is not present in the list or not pick an object at all. Furthermore, we allow the LLM to change the name of an action or delete an action and its associated entities entirely if it does not fit the context. Empirically we found that adding explicit instructions coupled with a few simple examples of what we expect from it, helps the LLM to better understand and solve the task at hand.

Finally, to get a better understanding of the context we prompt a small vision language model\footnote{https://huggingface.co/vikhyatk/moondream2 last accessed on 2nd of July 2025} with the following instruction: "In what scene does the action take place? Simply name the scene with no further explanations. Use very few words, just like a classification task, e.g., classroom, park, football field, mountain trail, living room, street." and prepend the answer to the proto-language. This allows the LLM to better understand the context of the actions and objects and thus better ground the description in the real world.

\begin{figure}
\includegraphics[width=1.0\linewidth]{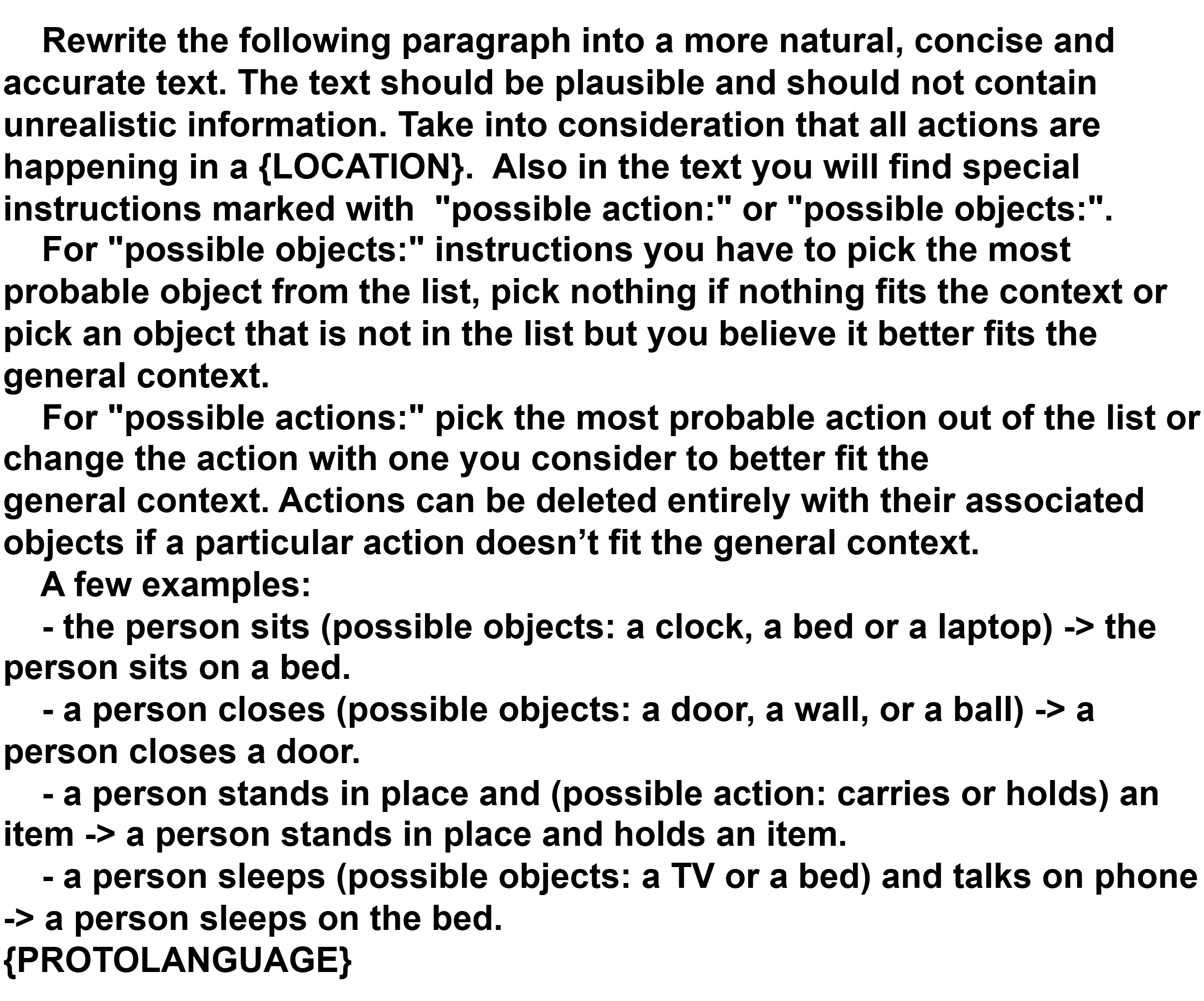}
\caption{The prompt used for generating the final text description.}
\label{fig:prompt}
\end{figure}

We concatenate the context information, special instructions and examples into the prompt used to generate the final text description, as depicted in Figure~\ref{fig:prompt}.

\section{Experimental Analysis}
\label{sec:experimental_chapter}
In this section, we describe the experimental settings, ranging from methods to datasets employed and the selected evaluation methodology.

\subsection{Datasets}
\label{sec:datasets}
To validate our approach, we employ five different datasets: Videos-to-Paragraphs~\cite{bogolin2020hierarchical}, COIN~\cite{tang2019coin}, WebVid~\cite{bain2021frozen}, VidOR~\cite{shang2019annotating} and ImageNet-VidVRD~\cite{shang2017video}.

\begin{figure}
\includegraphics[width=1.0\linewidth]{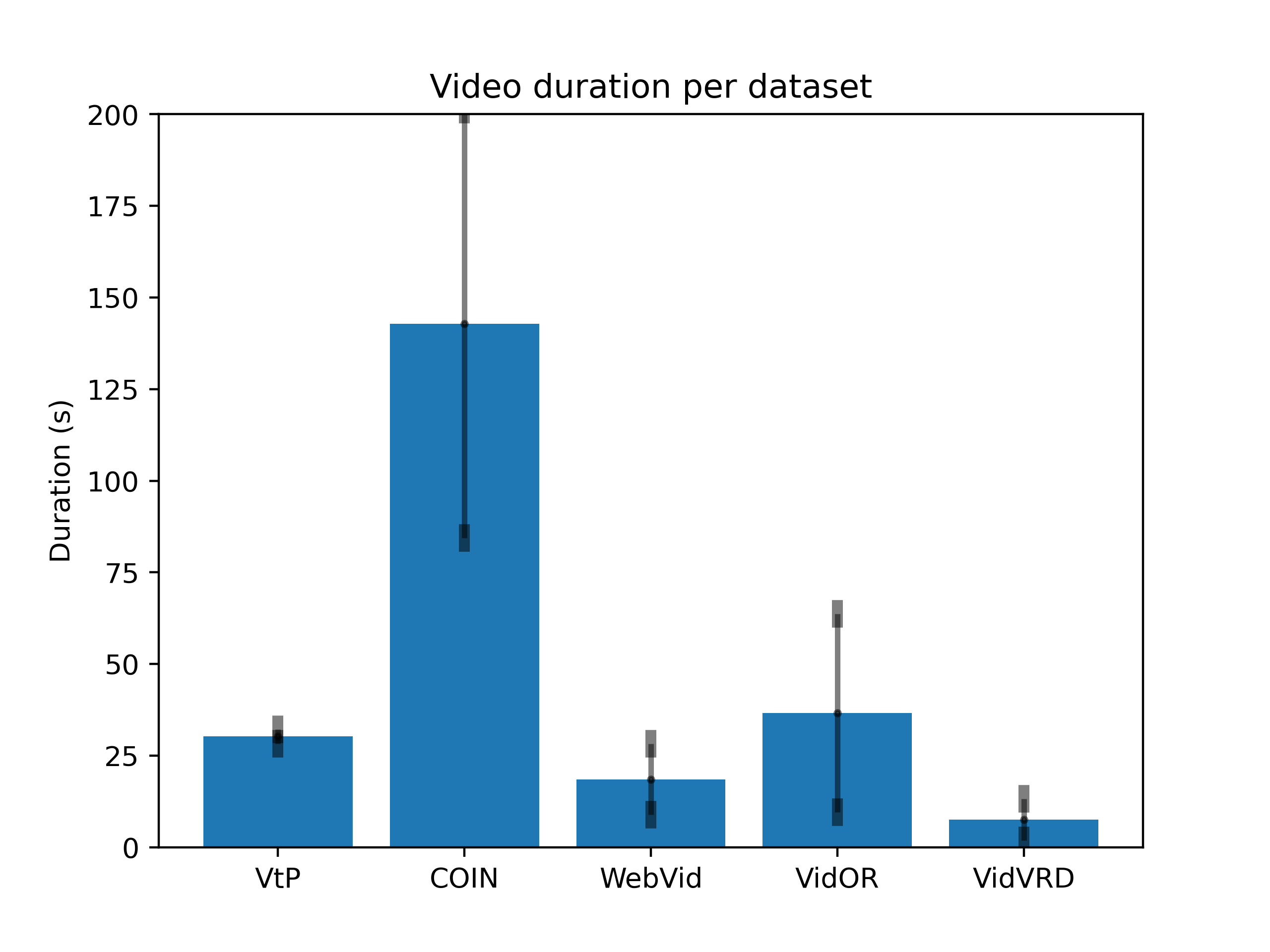}
\caption{Video average duration and its standard deviation per dataset. Note the diversity of the considered datasets, with regards to duration. VtP - Videos-to-Paragraphs.}
\label{fig:video_length}
\end{figure}

\textbf{Videos-to-Paragraphs} dataset ~\cite{bogolin2020hierarchical} consists of 510 videos of actions performed by actors in a school environment, filmed with both moving and fixed cameras. Videos contain varied and complex actions and activities, including interactions between two or more actors.  
\textbf{COIN} dataset~\cite{tang2019coin} has over 11000 videos of people solving 180 different everyday tasks in 12 domains (e.g., automotive, home repairs). Videos were collected from Youtube, with a relatively long duration of 2.36 minutes, which is why we chose this dataset.
\textbf{VidVRD}~\cite{shang2017video} dataset has 1000 videos, with 35 categories of annotated visual subject/object relations and a 132 predicate categories.
\textbf{VidOR} dataset has 11000 videos with 80 categories of objects and 50 categories of relations. Videos from both VidVRD and VidOR display rich visual relations between actors, who perform a multitude of complex intertwined actions. 
\textbf{WebVid} dataset~\cite{bain2021frozen} contains 10 million rich and diverse web-scraped videos with short text descriptions and a wide range of possible actions and environments. For each dataset, video duration statistics are presented in Figure~\ref{fig:video_length}.

\begin{figure}
\includegraphics[width=1.0\linewidth]{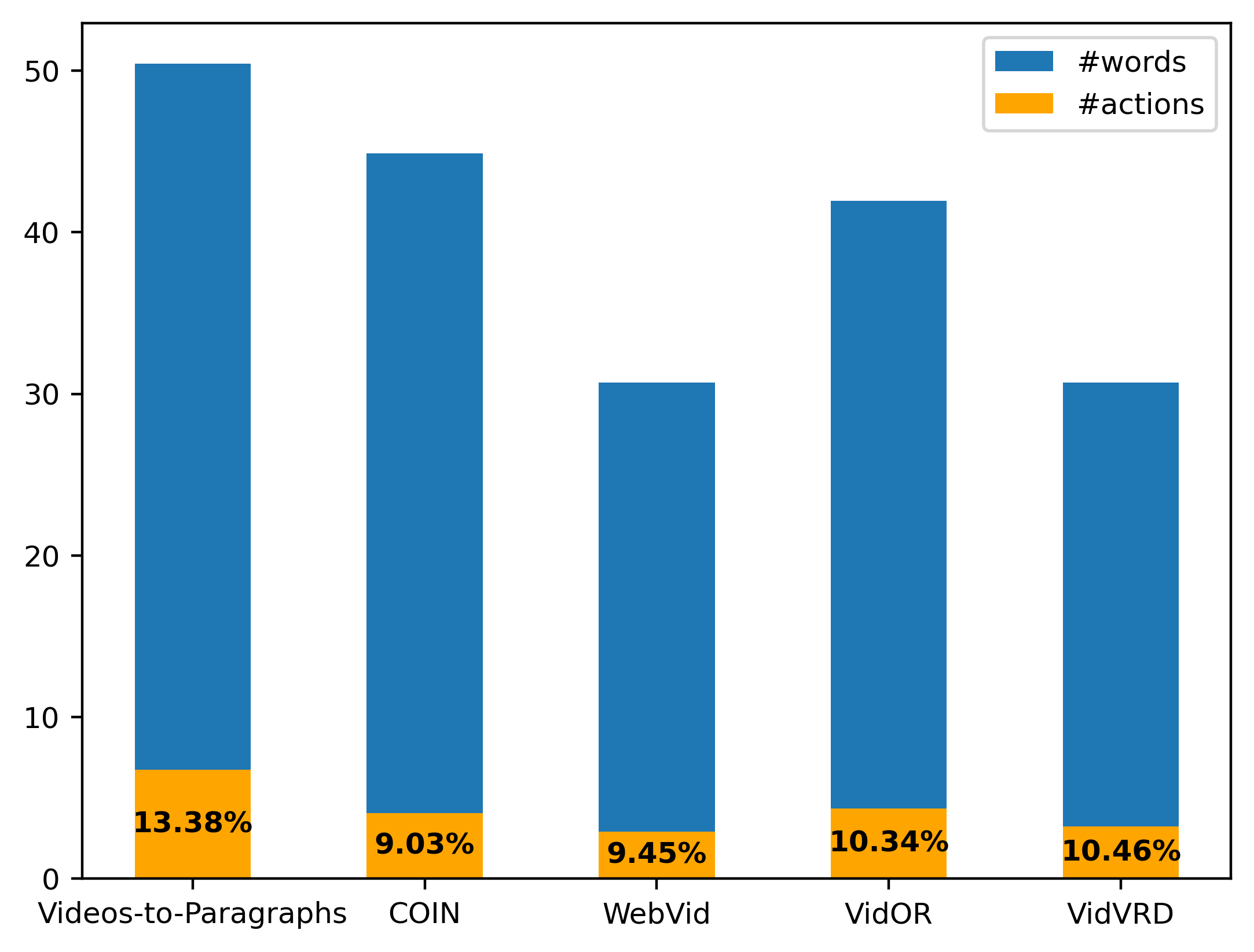}
\caption{Mean number of words, actions and percentage of actions over number of words in textual descriptions. Note that video duration is not necessarily a good measure of video complexity, with Video-to-Paragraphs video being the most complex.}
\label{fig:dataset_words_actions}
\end{figure}

It is important to analyze the different datasets based on their content and type of annotation.  
COIN dataset contains instructional videos that are significantly longer but could be accurately described by a specific temporal sequence of non-overlapping events and could also have a clear single global activity name (e.g., teaching how to install parquet). From this point of view, even though COIN videos are long, they are often simpler than videos in other datasets, such as the Videos-to-Paragraphs dataset, in which videos naturally depict complex activities that can take place in a classroom, but which cannot be correctly labeled by a single activity name. In fact, except for Videos-to-Paragraphs, all the other datasets contain actions that can be put into a limited number of well-defined labels. This brings them closer to the task of category recognition, rather than that of describing videos in free natural language. 

Videos-to-Paragraphs distinguishes itself through a hierarchical, two-level annotation. At the lower level, the activities are annotated by humans as a set of possibly overlapping subject-verb-object triplets. At the higher level, a single rich, paragraph-like description in more natural language is given by each human annotator, for each video. Videos in this dataset cannot be put into a single encompassing category. They are closer to the natural, everyday life, where multiple people can perform simultaneously multiple actions.

We analyzed the content of videos in the different datasets, as shown in Figure~\ref{fig:dataset_words_actions}, by counting the number of different actions and words averaged over each winning description per video, as selected by human ranking (see more details on human evaluation in Section~\ref{sec:evaluation}). To count the number of actions in a video, we consider only verbs that are the root of their respective sentences or are connected to the root via a \textit{CONJ} or \textit{ADVCL} dependency. As described by the different methods evaluated, it turns out that the 
Video-to-Paragraphs winning textual descriptions (as ranked by humans) contain the largest number of action, also having the highest percentage of actions per text length (more than 13\%). 

These observations lead to the conclusion that Videos-to-Paragraphs dataset is probably the most suitable for the task we are focusing in this paper: that of describing videos in richer, story-like language. The experiments also indicate a statistical difference between evaluating on this dataset vs. the others. Moreover, being a less-known dataset, it is almost sure that none of the considered methods have been trained on it. This is not the case for some of the other datasets: for example, VALOR was trained on a combination of datasets, which also contain WebVid videos. 

Our analysis leads to several \textbf{key observations about datasets:}

\begin{itemize}
    \item Video length is not necessarily a good proxy for video complexity as longer videos do not always contain more verbs and actions. We observe and advance a clear distinction between "general" video datasets and Video-to-Paragraphs dataset, a dataset specifically built for rich descriptions.
    \item The key differences between the two categories of resources are represented by the number of main actions required to properly describe the video, a complexity stemming from the video itself, as opposed to stemming only from the language description.
    \item Most of the visual and language models these days are trained and overfitted to a particular style of videos and short captions, mainly due to the lack of resources - complex and rich videos and captions.
\end{itemize}

\subsection{Methods compared}
We compare our approach (GEST) against a suite of existing open models: VidIL~\cite{wang2022language}, VALOR~\cite{chen2023valor}, COSA~\cite{chen2023cosa}, VAST~\cite{chen2023vast}, GIT2~\cite{wang2022git}, mPLUG-2~\cite{Xu2023mPLUG2AM} and PDVC~\cite{wang2021end}. 

Upon careful inspection of generated texts, we found that VidIL generated texts tend to be rich, but contain a high degree of hallucinations, while descriptions generated by our method tend to miss certain relevant aspects; see Section~\ref{sec:observations} for more details. 

Grounding VidIL and vice versa, adding more details to our approach should increase the overall quality of the descriptions. Therefore we add the proto-language generated in the GEST pipeline to the inputs used by VidIL (e.g. frame captions, events) and re-run GPT-4o to generate a textual description. Thus, the only changes we apply to the GEST pipeline are simply concatenating the inputs used by VidIL with the generated proto-language and some minimal tweaks to the prompt to understand and process the new input.

\noindent \textbf{Key observations about methods:}

\begin{itemize}
    \item VALOR~\cite{chen2023valor}, COSA~\cite{chen2023cosa}, VAST~\cite{chen2023vast}, GIT2~\cite{wang2022git} and mPLUG-2~\cite{Xu2023mPLUG2AM} are trained end-to-end on short and simple video captions, thus are prone to overfitting on this style. Grounding in vision is performed implicitly through end-to-end statistical learning, but it lacks clear explainability. As a result, the process is often ineffective and lacks a strong foundational basis.
    \item VidIL~\cite{wang2022language} splits the task by first extracting descriptors from uniformly sampled frames (i.e., by using CLIP~\cite{radford2021learning} and BLIP~\cite{li2022blip}) that are further fed to a text-only LLM. No explicit training is performed, as CLIP and BLIP are pre-trained and for generating the textual description, in-context learning can be used on the text-base LLM. This alleviates the overfitting issue and offers some control over the resulting text description.
    \item Our proposed approach leverages well-established tasks in computer vision, such as semantic segmentation, which have been extensively studied. We integrate multiple representations into an explicit, structured Graph of Events in Space and Time, that is deeply grounded in vision. This representation serves as the foundation for generating an initial, rich text description in the form of a proto-language. Finally, this proto-language is fed to a text-only LLM to build a refined, more natural sounding, complex textual description.
\end{itemize}

\subsection{Evaluation metrics}

\label{sec:evaluation}
To evaluate our approach and compare it with existing models, we use three evaluation protocols. First we turn to a text-based evaluation based on standard text similarity metrics, akin to how captioning methods are traditionally evaluated. Second, we 
perform a study with a relatively large number of human evaluators to
obtain a ranking of the generated texts based on multiple criteria such as richness and factual correctness. Third, we use a jury formed by several commercially available state-of-the-art LLM models with Vision capabilities, to rank all the methods.

\subsubsection{Classical metrics for video captioning}

We use an extensive list of text similarity metrics, using n-gram based metrics such as BLEU~\cite{papineni-etal-2002-bleu}, METEOR~\cite{banerjee-lavie-2005-meteor} or ROUGE~\cite{lin-2004-rouge} and more complex representations such as CIDEr~\cite{vedantam2015cider} and SPICE~\cite{anderson2016spice}. BERT-based text generation metrics include BERTScore~\cite{zhang2019bertscore} and BLEURT~\cite{sellam-etal-2020-bleurt} a state-of-the-art learned metric.

\subsubsection{Human ranking of video descriptions}

The classical metrics mentioned above have well-known limitations, as they have difficulty in capturing the semantics of text, bein g more focused on form rather than meaning. In order to better compare generated texts w.r.t to video content
at a semantic level, while taking in consideration nuanced crieria such as richness of description vs. correctness, we use human evaluators to rank the generated descriptions.
This study is performed by a volunteer group of 30 first-year Master students, with computer science background (BSc.). 

Human annotators are shown, for each given video, a number of automatically generated texts by the different anonymized methods we compare. Their task is to rank the descriptions from best to worst based on both richness and factual correctness. For more details about the instructions given to the human annotators see Appendix~\ref{sec:ann_tool}. 

\subsubsection{VLM-as-a-Jury ranking}

On a larger scale, instead of using human annotators, we harness strong Vision Language Models (VLMs) as judges. We scale not only the number of videos, but we also include the GEST + VidIL method and implement the LLM-as-a-Jury~\cite{verga2024replacing} approach (VLM-as-a-Jury), such as Claude 3.5\footnote{claude-3-5-sonnet-20241022}, GPT 4o\footnote{gpt-4o-2024-11-20}~\cite{hurst2024gpt}, Gemini\footnote{gemini-1.5-flash}, and Qwen2\footnote{Qwen2-VL-72B-Instruct}~\cite{Qwen2VL}. Each model is prompted with 10 uniformly sampled frames from each video, the generated descriptions and the same set of instructions as humans received.

We also evaluate the few-shot performance of both GEST and VidIL approaches to better gauge the differences between the two methods. To ensure consistency, we prompt the same panel LLM judges with virtually the same instructions, asking them to rate (i.e. GEST is better, VidIL is better or there is a tie) the two generated texts on the same previously mentioned criteria. In addition, we conduct an ablation study to assess the importance and impact of various modules within our approach. 

\noindent \textbf{Key observations about evaluation types:}

\begin{itemize}
    \item The first direction employed in our evaluation methodology is represented by standard text generation metrics.
    \item Next, we use human annotators to rank the considered methods on criteria such as richness or factual correctness.
    \item For the same task of ranking generated texts, we employ a jury of four large VLMs. We show that out of all the considered metrics, the VLM-as-a-Jury approach has the highest agreement with the human annotators (over 80\%). From there, we scale both the number of methods and number of videos using VLM-as-a-Jury as a proxy for human preferences.
    \item Using the VLM-as-a-Jury approach we evaluate the in context learning performance of both our method and VidIL.
    \item We perform an ablation study to better understand the importance of each module in our approach.
    \item Finally, we investigate the proposed direct self-supervised learning pathway from video to description, proving its effectiveness. 
\end{itemize}

\section{Evaluation Results}
\label{sec:eval_chapter}
In this section we present both quantitative and qualitative results, including both human and automated ranking, correlation between human preferences and VLM-as-a-Jury approach. In addition, we showcase qualitative examples and highlight some patterns observed throughout a multitude of videos and generated descriptions. Next we present a comparison in a few-shot learning performance between GEST and VidIL, while concluding with an ablation study in order to better understand the impact of each component in the system.

\begin{figure*}
\includegraphics[width=1.0\linewidth]{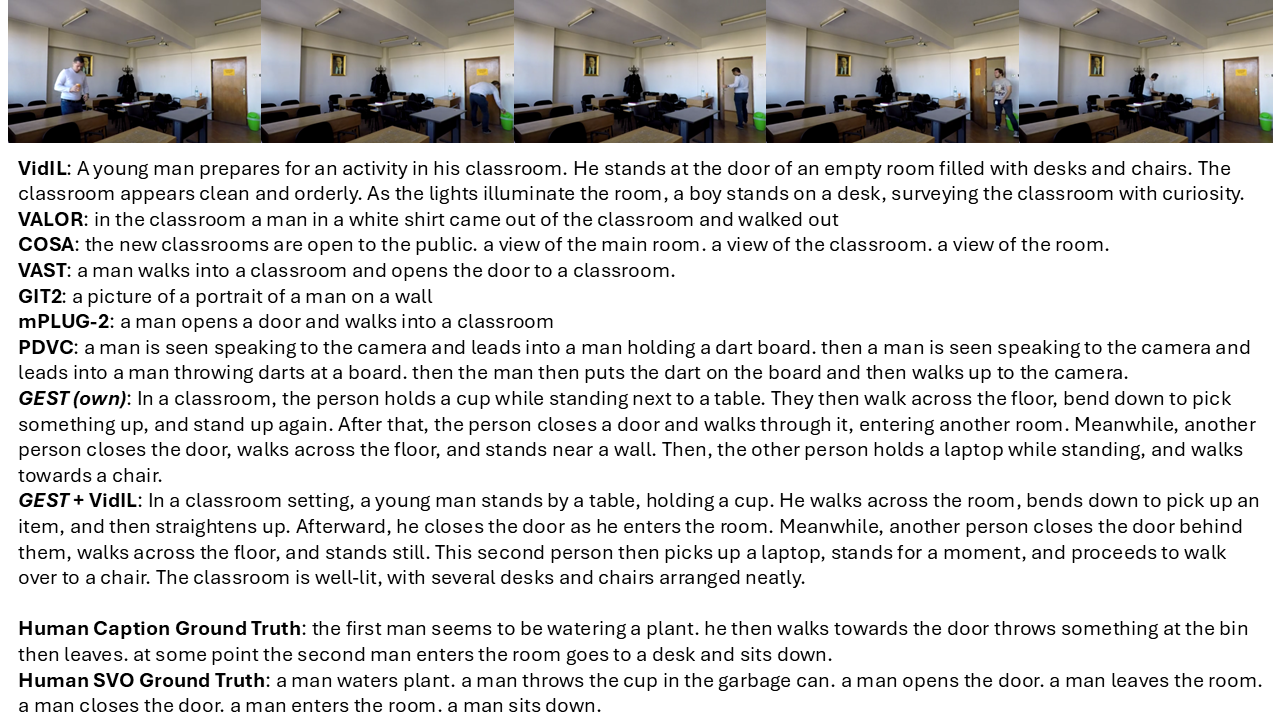}
\caption{Video descriptions generated by all considered methods together with ground truth for a video in Videos-to-Paragraphs dataset. Note that other methods completely miss the two different persons in the video, a crucial element. Most of the methods describe only the second person entering the room, completely missing the first person's actions. Our method correctly identifies that there are two distinct persons in the video and describes most of the actions in the videos, missing the first action due to the plant being out of frame and the last action due to the video ending prematurely.}
\label{fig:full_example_captions}
\end{figure*}

\subsection{Results with standard video captioning metrics}

Videos-to-Paragraphs dataset is the only one that offers rich ground truth paragraph-level descriptions, so we first evaluated all methods with respect to the human annotations using standard text generation metrics, which are presented in Table~\ref{tab:imar-caption-longest}. Results prove the overall quality of GEST, both on its own (second best) and in conjunction with VidIL (best).

\begin{table*}
  \centering
  \begin{tabular}{lc|ccccccc}
    \toprule
    Method & Avg  & B@4 & METEOR & ROUGE-L & CIDEr & SPICE & BERTScore & BLEURT \\
    \midrule
    VidIL~\cite{wang2022language} & 13.24 & 0.76&	9.95&	18.72&	1.69&	10.08&	10.99&	40.50\\
    VALOR~\cite{chen2023valor} & 12.38 & 0.35 &	5.24&	16.02&	1.41	&11.89	&16.59	&35.16\\
    COSA~\cite{chen2023cosa} &11.45 & 1.16&	6.58&	20.19&	\textbf{3.56}&	7.76&	0.15&	40.75\\
    VAST~\cite{chen2023vast} & 14.88 & 0.58&	6.11&	18.89	&1.59&	\underline{13.73}&	\textbf{22.44}&	40.83\\
    GIT2~\cite{wang2022git} & 13.61 & 0.23&	4.83&	16.34&	1.54&	11.99&	\underline{20.35}&	39.96\\
    mPLUG-2~\cite{Xu2023mPLUG2AM} & 12.14 & 0.03&	3.65&	11.89&	0.42&	\textbf{14.07} &18.32	&36.59\\
    PDVC~\cite{wang2021end}& {14.18} & 1.14 &	10.92 &	\textbf{24.02} &	1.85 &	9.98 &	7.25 &	44.10\\
    \textit{GEST (own)} & \underline{15.05} & \underline{1.19}& \underline{12.06}&	20.76&	\underline{2.84}&	8.32&	15.71	&\underline{44.47}\\
    \midrule 
    \multirow{2}{*}{\shortstack[l]{GEST+\\VidIL}} & \multirow{2}{*}{\textbf{16.30}} & \multirow{2}{*}{\textbf{1.84}} & \multirow{2}{*}{\textbf{14.32}} & \multirow{2}{*}{\underline{21.32}} & \multirow{2}{*}{1.68} & \multirow{2}{*}{13.15} & \multirow{2}{*}{16.61} & \multirow{2}{*}{\textbf{45.18}}\\
    & &\\
    \bottomrule
  \end{tabular}
  \caption{Videos-to-Paragraphs results when using the human annotated captions as ground truth. \textbf{Bold} marks the best result in each category, while \underline{underline} marks the second best. Note the strong performance of GEST, both in combination and standalone, as the top performer with competitive results on most of the considered metrics, being first or second on five of the seven metrics. B@4 stands for Bleu@4.}
  \label{tab:imar-caption-longest}
\end{table*}

\subsection{Results with human ranking}

As a first step in the annotation process, humans are faced with a test to ensure that they thoroughly read and understood the instructions. This test consists of asking the students to rank the three given descriptions of a video, for which the ranking is relatively simple based on common sense. We provide the video and the three descriptions in the supplementary material, for reference.
After filtering out students who failed to pass this test, students that annotated very few videos (less than 15) or who annotated videos surprisingly fast (e.g., in less time than the duration of the video), we are left with 15 annotators and 3889 annotated videos. We note that for this study, we use a subset of 811 videos (distributed as noted in Table~\ref{tab:eval-human-rank}), all annotated at least once, 810 videos out of the 811 have been annotated by at least two different users, while 799 videos have been annotated at least three times. 

In Table~\ref{tab:eval-human-rank} we present the results of the human annotation process. For Videos-to-Paragraphs dataset, GEST substantially outperforms the other models, with only one annotator rating it a close second (2.23 for mPLUG-2 vs 2.26 GEST), with the rest of the annotators rating it first. For the other datasets, we notice the very strong performance of VidIL, especially on COIN and WebVid. Although VidIL is still the preferred method for VidOR and VidVRD, the gap to other methods is smaller. Out of the considered methods, GIT2 and mPLUG-2 generate have by far the shortest and "simplest" descriptions (akin to video captioning; see examples in Figure~\ref{fig:full_example_captions}) and their similarity is clearly seen in the results: they are very close in human preference across all datasets. This is in somewhat contrast to the quantitative results, where the differences between the two methods are clearly visible. Somewhat similar, we notice a huge gap between the two evaluation protocols with PDVC: a competitive method if judged by text similarity metrics, but clearly underperforming if ranked by humans.

\begin{table*}
  \centering
  \begin{tabular}{lc|ccccc}
    \toprule
    \multirow{2}{*}{Method} & Average & VtP & COIN & WebVid & VidOR & VidVRD\\
    & & (489) & (75) & (92) & (93) & (62)\\
    \midrule
    VidIL~\cite{wang2022language} & \textbf{2.24} & 2.84 & \textbf{1.76} & \textbf{1.94} & \textbf{2.22} & \textbf{2.24}\\
    GIT2~\cite{wang2022git} & \underline{2.56} & \underline{2.73} & 2.49 & \underline{2.43} & \underline{2.51} & 2.61 \\
    mPLUG-2~\cite{Xu2023mPLUG2AM} & 2.60 & 2.84 & \underline{2.44} & 2.67 & 2.56 & \underline{2.51} \\
    PDVC~\cite{wang2021end} & 4.61 & 4.88 & 4.46 & 4.38 & 4.74 & 4.57\\
    \textit{GEST} & 2.99 & \textbf{1.71} & 2.85 & 3.22 & 3.33 & 2.86\\
    \bottomrule
  \end{tabular}
  \caption{Average rank (best is 1, worst is 5) as selected by human annotators. \textbf{Bold} marks the best result in each category, while \underline{underline} marks the second best. Note the very strong performance of GEST on Videos-to-Paragraphs dataset and VidIL for the other datasets. VtP - Videos-to-Paragraphs, numbers in parentheses indicate the number of videos included.}
  \label{tab:eval-human-rank}
\end{table*}

\subsection{Human ranking vs other metrics}
\label{sec:human_vs_metrics}

How well do different metrics correlate with the human preferences for the video description ranking task, and which one represents the best proxy for human annotator? Given two rankings, we define the agreement between them as the number of agreements at the pairwise rank ordering level: for each combination of two elements (descriptions), we check if they are rated in the same relative order in both rankings. Finally, we normalize this number by dividing it to the total number of combinations of two elements ($=$ total number of pairs) to generate the agreement score. 

In our experiments, we include the previously used text generation metrics together with the VLM-as-a-Judge and VLM-as-a-Jury approaches and present the results in Table~\ref{tab:metrics_agreement}. Out of the text similarity metrics, we note that BERTScore and SPICE metrics have the highest agreement with human preferences with a value of over 60, while Bleu@4 and BLEURT have a similar agreement of around 50, with METEOR and ROUGE-L having the lowest agreement. 

On the other hand, the VLM-as-a-Jury approach exhibits by far the highest agreement on each dataset with agreements of more than 77. The consistently high agreement between human preferences and VLM-as-a-Jury approach validates our evaluation protocol (i.e., the use of large VLMs as a proxy for human annotations). Furthermore, the "jury" exhibits a stronger agreement than each individual judge, manifesting a higher level of generalization than the individual large VLMs.

It is well known~\cite{reiter2018structured,reiter2009investigation,schmidtova2024automatic} that traditional metrics have fundamental limitations, and do not always correlate with human judgment, especially outside of their initial scope (e.g. machine translation for BLEU). Our experiments and results further emphasize these limitations and show that when human annotations are not available, a much better alternative for this task is represented by the VLM-as-a-Jury framework.

\begin{table*}
  \centering
  \begin{tabular}{lc|ccccc}
    \toprule
    \multirow{2}{*}{Method} & Average & VtP & COIN & WebVid & VidOR & VidVRD\\
    & & (489) & (75) & (92) & (93) & (62)\\
    \midrule
    \textit{Standard metrics}\\
    Bleu@4 & 51.27 & 51.25 & 53.20 & 53.91 & 51.40 & 46.61\\
    METEOR & 47.35 & 51.78 & 42.13 & 50.33 & 74.20 & 45.32 \\
    ROUGE-L & 43.36 & 45.79 & 38.53 & 46.74 & 43.66 & 42.40\\
    SPICE & 64.42 & 56.22 & 70.13 & 69.24 & 66.34 & 60.16\\
    BERTScore & 69.69 & 53.42 & 74.00 & 76.74 & 71.72 & 72.58\\
    BLEURT & 49.72 & 54.58 & 40.53 & 53.48 & 50.65 & 49.35\\
    \midrule
    \textit{Large VLMs}\\
    GPT4o & 76.25 & 84.17 & 73.33 & 76.96 & 70.97 & 75.81\\
    Claude & 77.12 & 81.47 & 78.67 & 76.81 & 73.01 & 75.65\\
    Gemini & 79.35 & 85.14 & 78.11 & \textbf{81.41} & 75.49 & 76.61\\
    Qwen2 & 74.47 & 81.29 & 69.07 & 77.83 & 71.72 & 72.42\\
    \midrule
    VLM-as-a-Jury & \textbf{80.79} & \textbf{86.13} & \textbf{80.67} & 80.98 & \textbf{77.63} & \textbf{78.55}\\
    \bottomrule
  \end{tabular}
  \caption{Agreement (range 0-100) between human preferences and considered metrics. \textbf{Bold} marks the highest agreement in each category. Out of the considered metrics, the VLM-as-a-Jury approach is the best proxy for human preferences. VtP - Videos-to-Paragraphs, numbers in parentheses indicate the number of videos included.}
  \label{tab:metrics_agreement}
\end{table*}

\subsection{VLM-as-a-Jury Ranking}

Using the VLM-as-a-Jury approach, we increase the number of videos and add the combination of GEST with VidIL and present the results in Table~\ref{tab:eval-vllm-rank-6}. The combination of GEST + VidIL is clearly rated the best on average, while being the top performer on all considered datasets. GEST retains its competitive performance on Videos-to-Paragraph, while being the second best model on average, leading the group of close performers formed together VidIL, GIT2 and mPLUG-2.

\begin{table*}
  \centering
  \begin{tabular}{lc|ccccc}
    \toprule
    \multirow{2}{*}{Method} & Average & VtP & COIN & WebVid & VidOR & VidVRD\\
    & & (489) & (318) & (443) & (478) & (164)\\
    \midrule
    VidIL~\cite{wang2022language} & 3.21 & 4.00 & \underline{2.82} & \underline{2.93} & 3.21 & \underline{3.11}\\
    GIT2~\cite{wang2022git} & 3.58 & 3.79 & 3.39 & 3.61 & 3.57 & 3.55\\
    mPLUG-2~\cite{Xu2023mPLUG2AM} & 3.53 & 3.85 & 3.11 & 3.63 & 3.63 & 3.44\\
    PDVC~\cite{wang2021end} & 5.50 & 5.77 & 5.15 & 5.65 & 5.39 & 5.53\\
    \textit{GEST} & \underline{3.16} & \underline{1.96} & 3.79 & 3.33 & \underline{3.16} & 3.55\\
    \midrule
    \textit{GEST} + VidIL & \textbf{2.02} & \textbf{1.64} & \textbf{2.74} & \textbf{1.86} & \textbf{2.03} & \textbf{1.84}\\
    \bottomrule
  \end{tabular}
  \caption{Average rank (best is 1, worst is 6) as selected by the VLM jury. \textbf{Bold} marks the best result in each category, while \underline{underline} marks the second best. The top performing method as evaluated using the VLM-as-a-Jury approach is the combination between GEST and VidIL. VtP - Videos-to-Paragraphs, numbers in parentheses indicate the number of videos included.}
  \label{tab:eval-vllm-rank-6}
\end{table*}

\subsection{Examples and Observations}
\label{sec:observations}

We present a sample video together with all the generated descriptions and ground truth in Figure~\ref{fig:full_example_captions}. By manually investigating more than 200 videos with their associated descriptions we noticed some strong patterns: both GIT2 and mPLUG-2 method generated very short descriptions in the form of one sentence, mentioning a single entity (that could include more than one actor e.g., "two men") and a single action. These descriptions are very simple, trivially true (a sentence that only describes the surroundings, or a sentence that states that a person is somewhere) and most of the time completely miss actions and actors. This makes them suitable for videos which have a single overarching action. 

While arguably competitive based on text-similarity metrics, PDVC-generated descriptions are too raw and contain way too little information to be relevant in real-world scenario. This proves yet again that automatic evaluation based on text similarity metrics is not the be-all-end-all solution for evaluating video descriptions as our analysis casts a serious doubt on the effectiveness of such an approach. 

On the other side, descriptions generated by VidIL are far richer, in some cases too rich, containing a lot of hallucinations and untrue facts. For example, in most Videos-to-Paragraph samples, as it sees a person in a room with a chalkboard it automatically infers that that person is a teacher, even if the person is sitting alone in the room, at a desk, doing something completely unrelated to teaching. It even hallucinates non-existing students (for some reason always six students) that are attentive to this imagined teacher, even if in the entire video there is only one person. Also, if a person is holding or writing on a laptop, that person "becomes" a computer scientist and all of the subsequent actions are described through this new persona (e.g. writing on laptop becomes coding). 

As our method is based on an action recognizer that has a rather small and fixed set of possible actions, our generated descriptions could lack flexibility and sometimes exhibit a limited understanding of the world. They tend to describe lower-level actions, for example, mopping the flooring might be described by holding an object while walking around. This also explains the strong performance of our method on Videos-to-Paragraphs dataset as the videos complexity stems from the multitude actions and interactions between multiple actors, rather than from individual action complexity. However, these limitations are not intrinsc to the approach, but rather to the limited number of actions available to the action detector. Moreover, giving the proto-language output to a powerful LLM, which is given the possibility to choose from the list of existing action names or replace them with new ones, to better fit the context, extends to possibilities and accuracy of our approach.

Also, by enhancing our method with a more diverse set of actions and objects, or equivalently grounding the rich set of VidIL inputs with clear, concrete actions, both obtained simultaneously by combining GEST and VidIL, leads to better descriptions. Such descriptions are grounded in a more diverse set of actions and activities, contain fewer hallucinations and better describe the source video.

\subsection{Zero-shot vs. few-shot performance}

\begin{table*}
  \centering
  \begin{tabular}{lcc|ccccccc}
    \toprule
    Method & N & Avg  & B@4 & METEOR & ROUGE-L & CIDEr & SPICE & BERTScore & BLEURT \\
    \midrule
    VidIL~\cite{wang2022language} & \multirow{ 2}{*}{0} & 13.24 & 0.76 & 9.95	& 18.72	& 1.69	& 10.08 &	10.99& 	40.50\\
    \textit{GEST} &  & \textbf{15.05} & \textbf{1.19}&	\textbf{12.06}&	\textbf{20.76}&	\textbf{2.84}&	\textbf{8.32}&	\textbf{15.71}	&\textbf{44.47}\\
    \midrule
    VidIL & \multirow{ 2}{*}{1} & 14.91 & 1.38 & 10.72&	20.49	&2.03&	\textbf{11.53}&	15.08&	43.16 \\
    \textit{GEST}&  & \textbf{16.91} & \textbf{1.84}&	\textbf{12.17}&	\textbf{21.61}&	\textbf{4.33}&	10.04&\textbf{20.59}&	\textbf{45.02}\\
    \midrule
    VidIL& \multirow{ 2}{*}{3} & 15.92 & 1.93&	11.11&	21.64&	3.06&	\textbf{12.26}&	17.28&	44.19\\
    \textit{GEST} &  & \textbf{17.52} & \textbf{2.37}&	\textbf{12.11}&	\textbf{22.87}&	\textbf{4.98}&	11.38&	\textbf{23.54}&	\textbf{45.38}\\
    \midrule
    VidIL& \multirow{ 2}{*}{5} & 16.61 & 2.47&	\textbf{12.05}&	22.45&	2.42&	\textbf{13.16}&	18.64&	45.08\\
    \textit{GEST} &  & \textbf{17.62} &\textbf{2.50}&	11.86&	\textbf{22.99}&	\textbf{4.65}&	11.79&	\textbf{24.02}&	\textbf{45.53}\\
    \bottomrule
  \end{tabular}
  \caption{Videos-to-Paragraphs results in fewshot settings, using human annotated captions as ground truth. \textbf{Bold} marks the best result in each setting. The best average performance is obtained by GEST in five-shot setting, with GEST being the top performer in five out of the seven metrics. N stands for the number of solved examples (shots) provided to the model.}
  \label{tab:metrics-fewshot}
\end{table*}

\begin{figure*}
\includegraphics[width=1.0\linewidth]{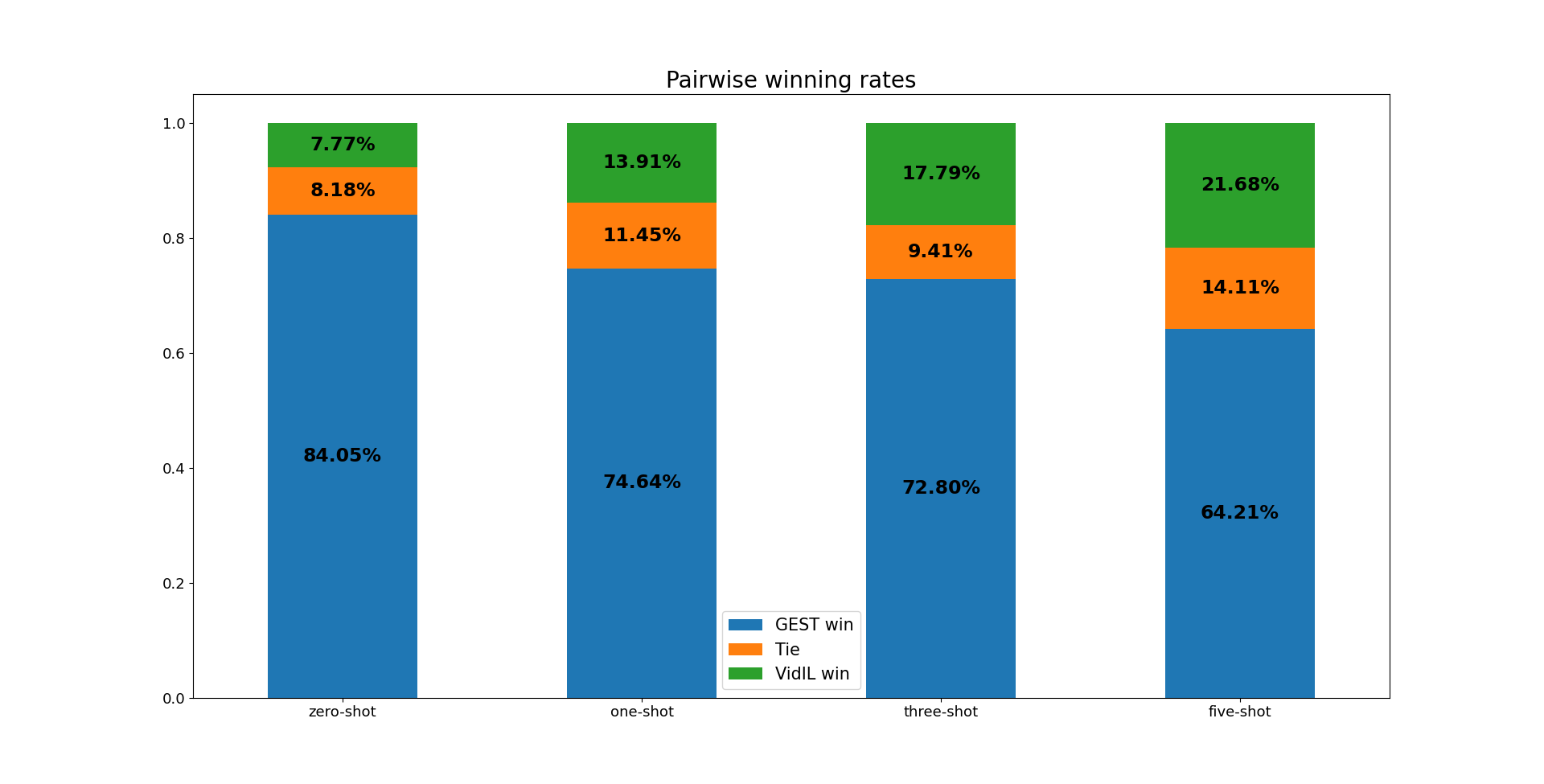}
\caption{Zero and few-shot pairwise winning rates as selected by VLM-as-a-Jury for videos in Videos-to-Paragraphs. Observe the dominant performance of GEST, with VIDIL adapting as the number of examples increases.}
\label{fig:few_shot}
\end{figure*}

Since the final stages of both our method and VidIL are based on LLMs, they are inherently influenced by advancements and techniques commonly employed in LLMs, such as In-Context Learning (ICL). Therefore we add one, three and five solved examples (using the human ground truth caption) and ask the LLM to solve the next one. Text similarity metrics for different fewshot setting are presented in Table~\ref{tab:metrics-fewshot}, while pairwise winning rates for zero, one, three and five-shot setting are presented in Figure~\ref{fig:few_shot}. As GEST is strongly grounded, it has a very strong zero-shot performance, winning in more than 84\% of cases. As we provide more and more examples (randomly sampled; using ground truth captions) the gap between GEST and VidIL shrinks, with GEST winning 64\% versus 21\% for VidIL. We note that the quality of descriptions generated by both VidIL and GEST increases as more and more examples are provided.

\subsection{Ablation study}

Our proposed method uses somewhat independent blocks (e.g., action detection, semantic segmentation) that work together to build a stronger, more faithful Graph of Events in Space and Time. In this section, we exclude some of these modules in order to better understand the importance of each component. We perform this ablation study on the Videos-to-Paragraphs dataset and evaluate using GPT-4o as a Judge. As the differences between the generated texts are subtle, we find that text similarity metrics (e.g., BLEU, ROUGE) are not powerful enough to capture such nuances (e.g., the total number of unique persons in a video). In Table~\ref{tab:ablation} we present the results of the ablation study in which we drop one module a time, starting with the semantic person re-identification, continuing with depth estimation and finally dropping the semantic segmentation module. As a recap, the semantic person re-identification is based on HSV histogram, and is used to re-identify persons that reappear (either due to tracking faults or due to exiting the frame). Depth estimation is used to filter objects that appear near a person in the frame (2D) but are actually far away (in the real 3D world). Finally, semantic segmentation was used to complement and refine the objects found by the object detector. 

Turning to Table~\ref{tab:ablation} we note that the most important component is the semantic segmentation: without using it, the average ranking drops from 1.57 all the way to 3.66 (with 4 being the absolute worst). Dropping the person re-identification module greatly increases the number of unique persons in the videos (e.g., if a person exists the frame and re-enters later and at another position it would be identified as a "new" person), dropping the overall quality of the graph and consequently of the generated description. Finally, dropping the depth estimation module makes the proto-language longer (i.e., with more possible interactions for each event) and the task for the "refining" LLM harder by adding objects that are not in the proximity of the person or persons involved in given event.

\begin{table}
  \centering
  \begin{tabular}{lc}
    \toprule
    \multirow{2}{*}{Method} & VtP\\
    & (489)\\
    \midrule
    GEST & 1.57 \\
    w/o semantic person re-id & 2.47 \\
    w/o depth estimation & 2.36 \\
    w/o semantic segmentation & 3.66\\
    \bottomrule
  \end{tabular}
  \caption{Average rank (best is 1, worst is 4) as selected by GPT-4o Judge on Videos-to-Paragraphs (VtP) dataset. Note the huge drop of performance when not using the semantic segmentation task. }
  \label{tab:ablation}
\end{table}

\subsection{Self-supervised end-to-end learning
pathway from video to description}

In this Section we set to investigate how our proposed modular, neuro-analytic pathway, can be used as an automatic teacher, in a
self-supervised regime, to train a direct, end-to-end neural student pathway. The self-supervised stage could function as pre-training to help boost the performance of existing methods. As such we finetune video captioning methods on Videos-to-Paragraphs dataset, with and without using machine-generated video descriptions. We use GEST-generated textual descriptions for videos in Videos-to-Paragraphs dataset in an additional self-supervised manner with GEST teacher and neural path student step , followed by a standard finetuning step using ground truth captions. 

\begin{figure*}
\includegraphics[width=1.0\linewidth]{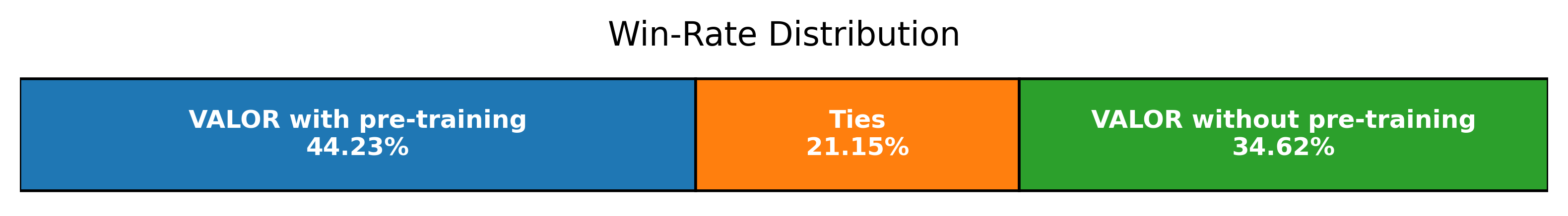}
\caption{Head-to-head comparison to investigate the effect of pre-training (self-supervised with GEST teacher) on generated data, judged using VLM-as-a-Jury approach. Our self-supervised GEST pre-training on generated data increase the overall quality of the video descriptions.}
\label{fig:valor_winrate}
\end{figure*}

In Figure~\ref{fig:valor_winrate} we present the win-rate (via the same VLM-as-a-Jury approach) obtained by comparing the performance of directly finetuning on the target dataset with the model that is further pre-trained and then finetuned. The same experiment with evaluation performed using text generation metrics is presents in Table~\ref{tab:end_to_end_valor}. Under both evaluation protocols, we observe that pre-training on generated descriptions generates an important and constant improvement. This proves that our method generates qualitative and informative descriptions, descriptions that can be used to further improve existing video captioning methods. 

These promising results open up a new avenue for improving existing video captioning methods, via increasing the size and quality of existing resources. Besides just mentioned video-descriptions pairs, using our approach we have generated video-descriptions tuples (i.e. for each video we have 6 descriptions that are ordered by quality, either by human annotators or large VLMs) that can readily be used for training vision language models (e.g. via Direct Preference Optimization~\cite{rafailov2023direct}).

\begin{table*}[h!]
  \centering
  \begin{tabular}{lc|ccccccc}
    \toprule
    Method & Avg  & B@4 & METEOR & ROUGE-L & CIDEr & SPICE & BERTScore & BLEURT \\
    \midrule
    VALOR & 26.68 & 14.27 & 14.40 & 35.34 & 21.51 & 21.70 & 32.31 & 47.24\\
    VALOR$^*$ & \textbf{28.92} & \textbf{15.61} & \textbf{15.54} & \textbf{36.50} & \textbf{27.87} & \textbf{22.16} & \textbf{36.82} & \textbf{47.95}\\
    \bottomrule
  \end{tabular}
  \caption{Finetune results on Videos-to-Paragraphs test set. \textbf{Bold} marks the best result in each category. $^*$ marks models that are pre-trained, with our self-supervised teacher-student approach, on descriptions generated by our GEST method, as an initial step. Note that the self-supervised pre-training increases the performance on all considered metrics, proving the effectiveness of the proposed approach.}
  \label{tab:end_to_end_valor}
\end{table*}

\subsection{Final remarks on experiments}

\begin{figure}[!h]
\includegraphics[width=1.0\linewidth]{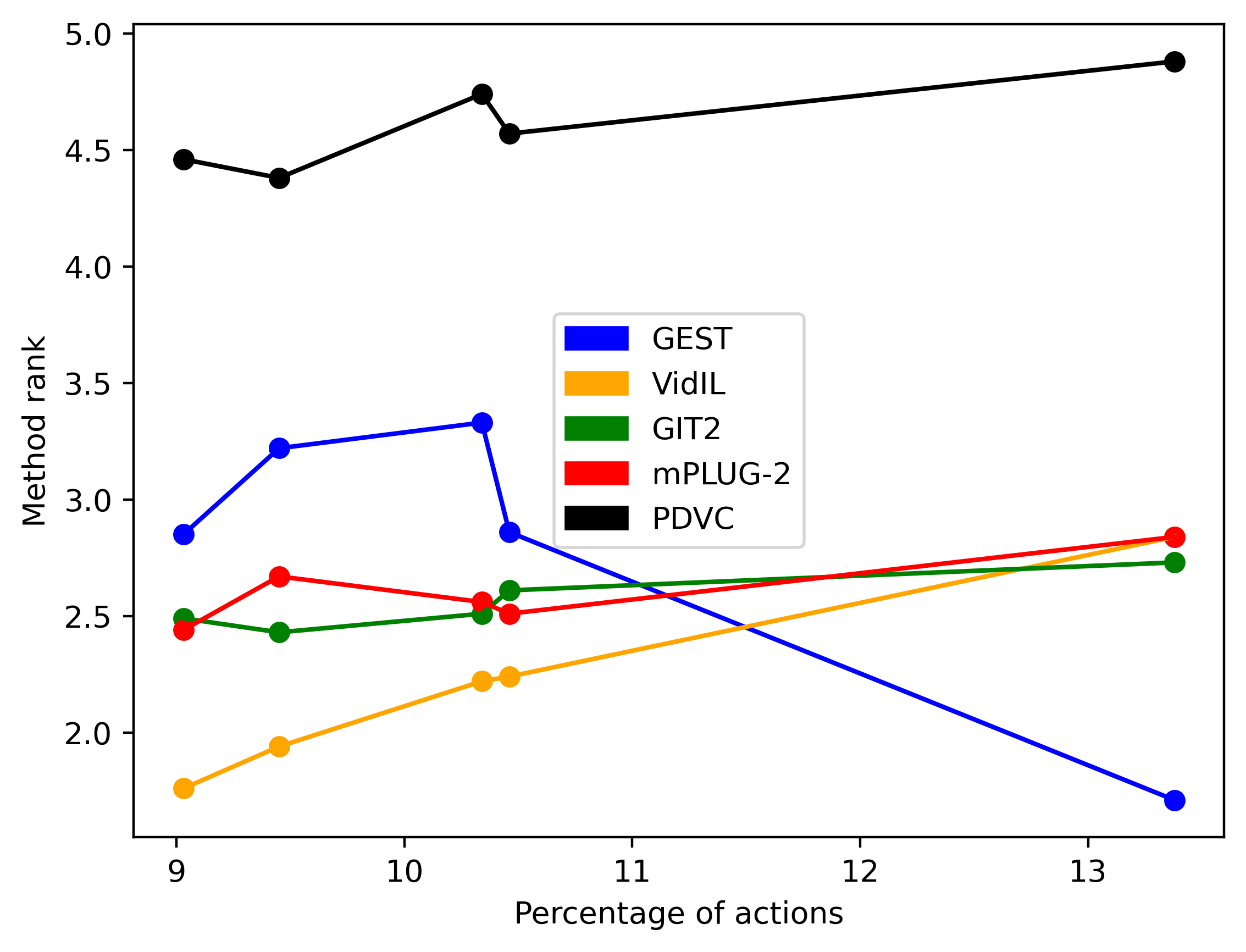}
\caption{Average rank as given by human (Y axis) versus the percentage of main actions for each dataset (X axis). GEST is the only method that becomes better as the complexity of the videos increases.}
\label{fig:ranks_vs_verbs_human}
\end{figure}

Our studies and experiments show how existing datasets and consequently models, present fundamental differences. The two categories identified, namely resources for simpler, shorter video captions compared to more complex videos and rich description, are fundamentally and statistically different. In Figure~\ref{fig:ranks_vs_verbs_human} we plot the average rank as given by the human annotators for each method together with the percentage of main action for each dataset. 

On datasets with richer, more complex videos such as Videos-to-Paragraphs, our method consistently outperforms other methods under all evaluations types: using text generation metrics, human annotators and VLM-as-a-Jury approach. Results across evaluation dimension are presented in Table~\ref{tab:vtp_across_evaluation} and prove the effectiveness of our method for this type of datasets.

Finally, we show that the generated descriptions are informative and can be used for improving existing methods (e.g. via direct pre-training).

\begin{table}[h!]
  \centering
  \begin{tabular}{lccc}
    \toprule
    Method & Text Metrics & Humans & VLM Jury\\
    \midrule
    VidIL~\cite{wang2022language} & 4 & 3 & 4 \\
    GIT2~\cite{wang2022git} & 3 & 2 & 2\\
    mPLUG-2~\cite{Xu2023mPLUG2AM} & 2 & 3 & 3 \\
    PDVC~\cite{wang2021end} & 5 & 5 & 5\\
    \textit{GEST} & 1 & 1 & 1\\
    \bottomrule
  \end{tabular}
  \caption{Absolute ordering (best is 1, worst is 5) of the methods under different evaluations (i.e. text generation metrics, human evaluation and VLM-as-a-Jury) on the Videos-to-Paragraphs dataset. Under all evaluation directions, from humans to text generation metrics and VLM jury, GEST is the top performer.}
  \label{tab:vtp_across_evaluation}
\end{table}

\section{Conclusions}

We proposed a novel method that combines state-of-the-art models from both computer vision and natural language processing domains with a procedural module to generate rich, explainable video descriptions. It uses object and action detectors, semantic segmentation and depth estimation to automatically extract frame-level information, that is further aggregated into video-level events, ordered in space and time. Using a modular algorithm in which multiple learnable tasks are connected in an explainable way, events and their spatio-temporal relations are further converted into a GEST-based proto-language that is both trustworthy and rich in information. Using LLMs at a final stage, the GEST-based proto-language is finally converted into a fluent and coherent story, which describes the events in natural language. To our best knowledge, we are the first to explore such a procedural multi-task approach that offers both a shared understanding between vision and language, in the form of graphs of events in space and time, and a self-supervised learning scheme in which a direct, fully neural student path is trained on the output of the more analytical GEST teacher path. Our extensive experiments and analysis, on several datasets with several types of evaluation, both human and automatic, provide a solid validation for our approach. They could also constitute a solid base for future research in the exciting world of vision-language.

Finally, our approach could open a new direction - a foundation that could lead to improvements of large VLM models. It aims to achieve this through an explainable, multi-task model that clearly defines each step from the raw visual input to sophisticated language output. 

\bmhead{Acknowledgements}

This work is supported in part by projects “Romanian Hub for Artificial Intelligence - HRIA”, Smart Growth, Digitization and Financial Instruments Program, 2021-2027, MySMIS no. 334906 and EU Horizon project ELIAS (Grant No. 101120237).


\begin{appendices}

\section{Guidelines and Annotation Process}
\label{sec:ann_tool}

Besides defining the task, we also provide general guidelines to aid in the annotation process and properly describe the rich video description task. As such we ask users to rank videos on three directions, in order of priority: description richness (we want description to be semantically rich, we are not particularly interested in blunt, trivially true statements), factual correctness (we don't want descriptions to be rich for the sole purpose of being rich; they must be grounded in reality). Finally, we prefer grammatically correct sentences. 

\begin{figure*}
\includegraphics[width=1.0\linewidth]{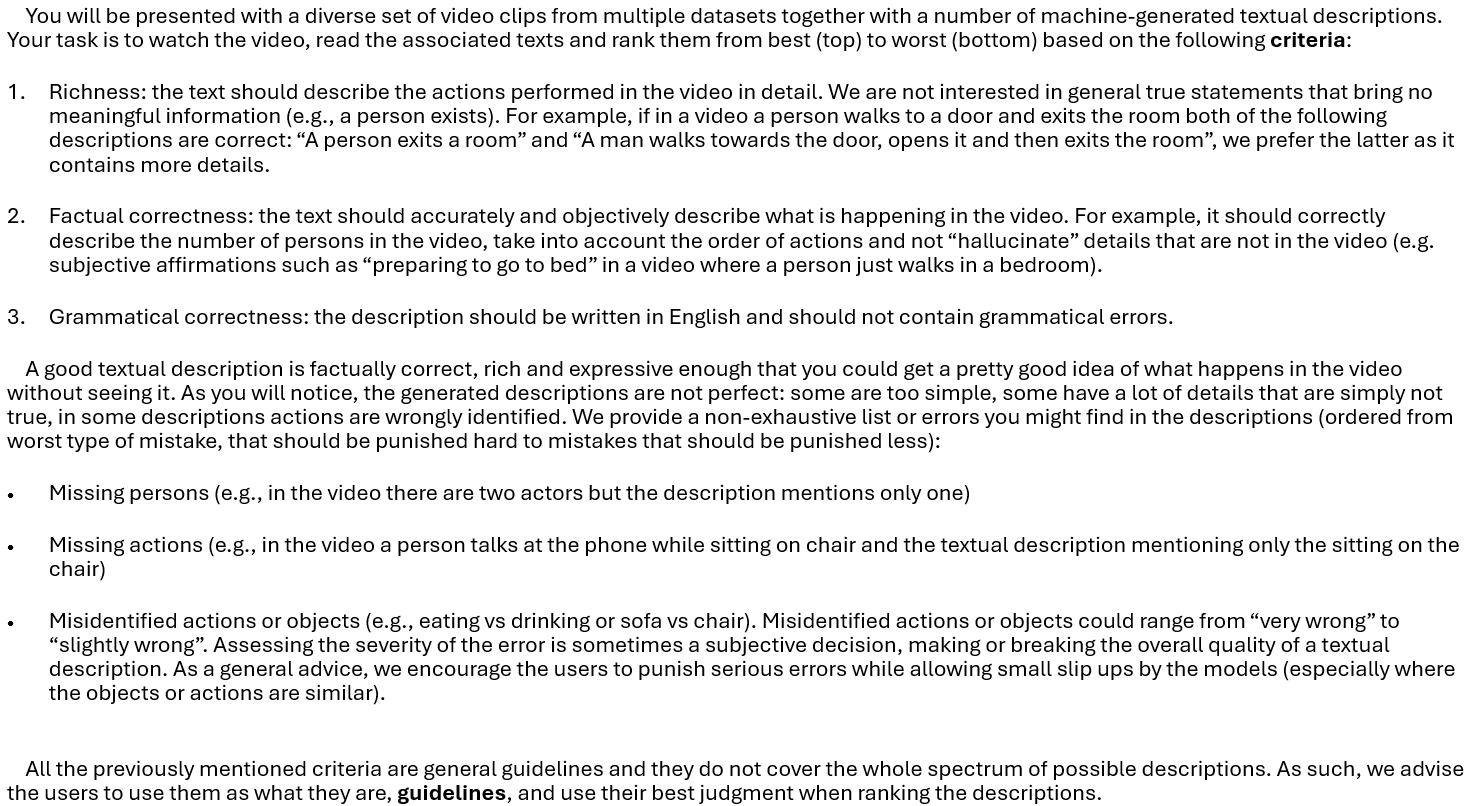}
\caption{Annotation instructions given to both humans and VLMs.}
\label{fig:instructions}
\end{figure*}

As the descriptions are not always perfect, we provide users with a few general rules as to what kind of errors to punish more than others. We ask them to punish descriptions that have the wrong number of actors more severely, followed by descriptions that miss actions. Finally, we consider misidentified actions or objects to be less of a problem. The annotation guidelines given to humans and large VLMs are presented in Figure~\ref{fig:instructions}.

\begin{figure*}
\includegraphics[width=1.0\linewidth]{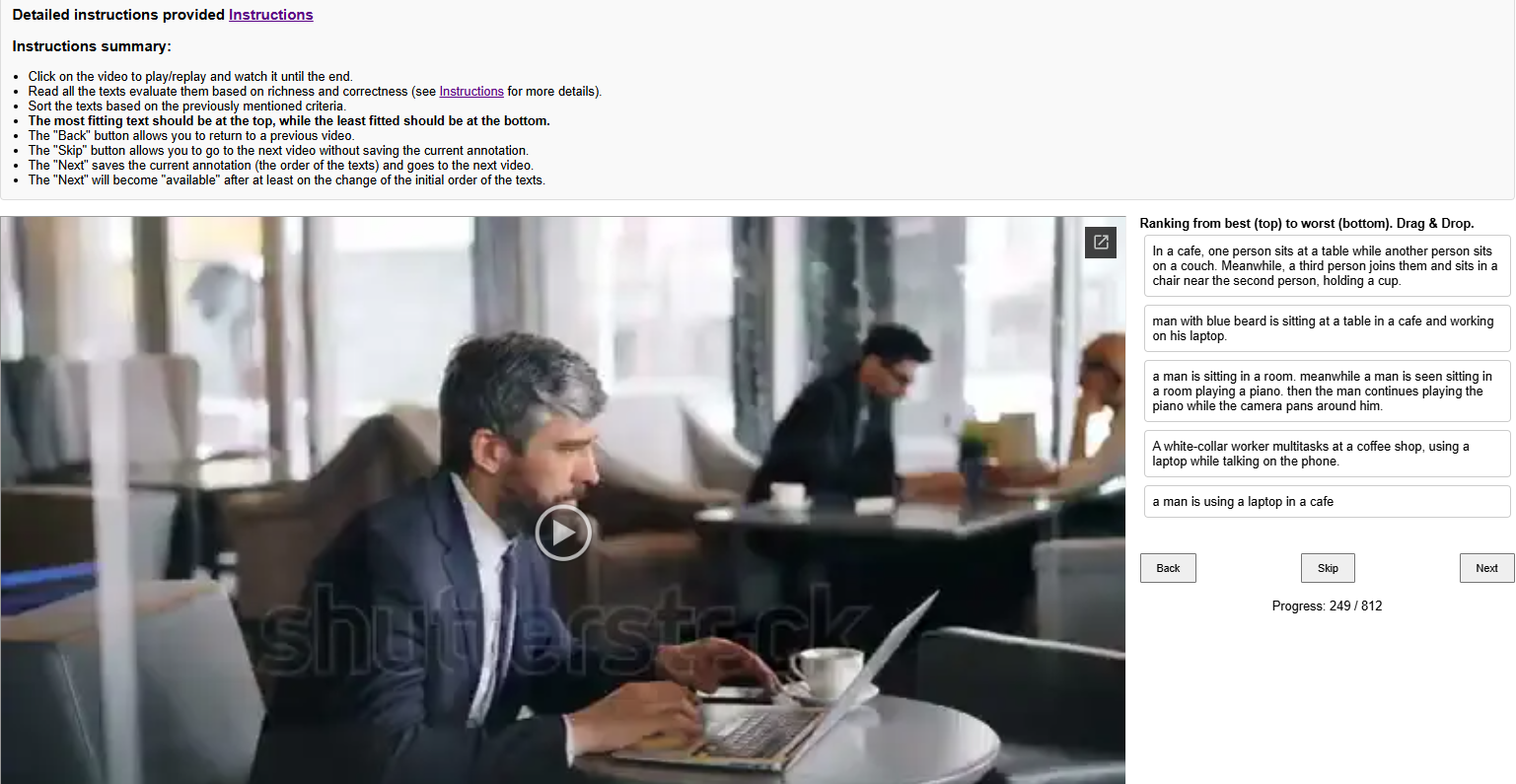}
\caption{Screenshot of the annotation tool, containing in the top part a short list of instructions, including a link to the full annotation guidelines. On the bottom part, a large viewport for the video, with texts presented in the right part.}
\label{fig:ann_tool}
\end{figure*}

We kept a simple interface for the annotation tool, with a reminder and a link to the general guidelines, a large viewport for the video, with the five rankable descriptions on the right. For each video, we randomize the order in which descriptions are presented to each user to avoid positional biases. We allow users to skip videos and return to the previous annotation if needed. A snapshot of the annotation platform is presented in Figure~\ref{fig:ann_tool}.

\end{appendices}


\bibliography{sn-bibliography}

\end{document}